# The GRT Planning System: Backward Heuristic Construction in Forward State-Space Planning

**Ioannis Refanidis**                                    YREFANID@CSD.AUTH.GR
**Ioannis Vlahavas**                                     VLAHAVAS@CSD.AUTH.GR
*Aristotle University*
*Dept. of Informatics*
*54006 Thessaloniki, Greece*

## Abstract

This paper presents GRT, a domain-independent heuristic planning system for STRIPS worlds. GRT solves problems in two phases. In the pre-processing phase, it estimates the distance between each fact and the goals of the problem, in a backward direction. Then, in the search phase, these estimates are used in order to further estimate the distance between each intermediate state and the goals, guiding so the search process in a forward direction and on a best-first basis. The paper presents the benefits from the adoption of opposite directions between the preprocessing and the search phases, discusses some difficulties that arise in the pre-processing phase and introduces techniques to cope with them. Moreover, it presents several methods of improving the efficiency of the heuristic, by enriching the representation and by reducing the size of the problem. Finally, a method of overcoming local optimal states, based on domain axioms, is proposed. According to it, difficult problems are decomposed into easier sub-problems that have to be solved sequentially. The performance results from various domains, including those of the recent planning competitions, show that GRT is among the fastest planners.

## 1. Introduction

So far, planning problems have been considered as a special kind of particularly difficult search problems (Newell & Simon, 1972) and many algorithms for decomposition, abstraction, least commitment etc. have been proposed to cope with them. In the early 90's, researchers were arguing that plan-space planning is more efficient than state-space planning (Barrett & Weld, 1994; McAllester & Rosenblitt, 1991; Minton, Bresina & Drummond, 1994; Penberthy & Weld, 1992). In the mid 90's, new algorithms appeared that achieved even better performance by transforming planning problems either into graph solving problems (Blum & Furst, 1995, 1997) or into satisfiability ones (Kautz & Selman, 1992, 1996, 1998). However, it has been shown that simple search strategies with the use of domain-dependent heuristics can solve large problems (Gupta & Nau, 1992; Korf & Taylor, 1996; Pearl, 1983; Slaney & Thiebaux, 1996).

In recent years, part of the planning community turned towards heuristic planning, adopting known search strategies and developing powerful domain-independent heuristics that achieve significant performance. The first planner was UNPOP (McDermott 1996, 1999) and was followed by ASP (Bonet, Loerings & Geffner, 1997), HSP (Bonet & Geffner, 1998), HSPr (Bonet & Geffner, 1999), GRT (Refanidis & Vlahavas, 1999b), FF (Hoffmann & Nebel, 2000) and ALTALT (Nigenda, Nguyen & Kambhampati, 2000). These domain independent heuristic planners search for solutions either in the state-space or in the regression space. Most of them use variations of a relatively simple idea as a guide: they estimate the distance between two states, based on estimates of the distances between each fact of the problem and one of the two states.

The above planners can primarily be classified based on the forward or backward direction, in which the heuristic is constructed and the state-space is traversed. We distinguish the following three categories:

- Forward heuristic construction, forward search (ASP, HSP, FF).
- Forward heuristic construction, backward search (HSPr, ALTALT).
- Backward heuristic construction, forward search (UNPOP, GRT).

Generally, the forward direction seems to be more advantageous than the backward one, both when constructing the heuristic and when searching, because in the backward direction and in case of incomplete goal states, problems with invalid states and unreachable facts usually arise. However, using the forward direction for both tasks requires reconstructing the heuristic function for each visited state, spending in this way a significant portion of the processing time, while using opposite directions for both tasks allows constructing the heuristic once, in a pre-processing phase.

This paper presents the GRT planning system. It is the only domain independent heuristic planner that constructs the heuristic once, in a backward direction and in a pre-processing phase. UNPOP, although it uses the same directions, reconstructs the heuristic from scratch for each visited state. GRT, in a pre-processing phase estimates the distance between each fact and the goals of the problem. During the search phase, these estimates are used in order to further estimate the distance between each visited state and the goals, guiding so the search process in a forward direction and on a best-first basis. Constructing the heuristic once offers the ability to evaluate states very quickly, while traversing the state-space in a forward direction allows the planner to avoid invalid states that arise in the regression space.

The paper substantially extends previous work (Refanidis & Vlahavas, 1999b, 1999c, 2000a and 2000b), in that it presents and proves the fundamental theory of the planner, along with many new techniques developed on it, it extensively tests the contribution of each technique to its overall performance and provides a thorough comparison to other planning systems.

The rest of the paper is organized as follows: Section 2 presents the data structures and the main algorithms of the planner. Section 3 discusses the difficulties that incomplete goal states cause to the backward direction of the construction of the heuristic and presents methods to cope with them. The same methods are also applied to identify and enrich poor domain representations.

Two approaches to reduce the problem's size are presented in Section 4. The first one deals with the identification and elimination of irrelevant objects and the second one concerns the adoption of a numerical representation of resources.

Section 5 deals with the problem of local optimal states and proposes a method to cope with them. Specifically, the XOR-constraints are introduced and used in order to decompose difficult problems into easier sub-problems that have to be solved sequentially. Section 6 presents the operation of GRT, Section 7 presents the related work and Section 8 presents performance results, which show that GRT is among the fastest domain-independent planners. Finally, Section 9 concludes the paper and poses future directions.

## 2. The GRT Heuristic

In STRIPS (Fikes & Nilsson, 1971), each action $a$ is represented by three sets of facts: the precondition list $Pre(a)$, the add-list $Add(a)$ and the delete-list $Del(a)$, where $Del(a) \subseteq Pre(a)$. A state $S$ is defined as a finite set of facts. An action $a$ is *applicable* to a state $S$ if:

$$Pre(a) \subseteq S \qquad (1)$$

The state resulting from the application of an action $a$ to state $S$ is defined as:

$$S' = res(S,a) = S \setminus Del(a) \cup Add(a) \tag{2}$$

Inductively we can define the state resulting from the application of a sequence of actions ($a_1$, $a_2$, ..., $a_N$) to a state $S$ as:

$$S' = res(S, (a_1, a_2, ..., a_N)) = res(\ res(S, (a_1, a_2, ..., a_{N-1})), a_N) \tag{3}$$

with the requirement that each action $a_i$ is applicable to the state $res(S, (a_1, a_2, ..., a_{i-1}))$, for each i=1, 2, ..., N. In the formalization used henceforth, the set of problem constants is assumed to be finite and no function symbols are used, so the set of actions is finite.

A planning problem $P$ is a triplet $P=(O, Initial, Goals)$, where $O$ is the set of ground actions, *Initial* is the initial state and *Goals* is a set of facts. The task is to find a sequence of actions $a_1$, $a_2$, ..., $a_N$ that can be applied to the initial state, so that the state resulting from their application will be a superset of *Goals*. The sequences of actions are called *Plans*. A plan that can be applied to the initial state is called a *valid plan*. A valid plan that achieves the *Goals* is called a *solution* of the planning problem. A planning problem may have several or no solutions. In the latter case the problem is described as *unsolvable*.

The next sub-section gives a brief presentation of the ASP heuristic, which was our motivation and helps to understand the following concepts, whereas the subsequent sub-sections present the GRT heuristic in detail.

## 2.1   The ASP Heuristic

In the ASP heuristic, for each action $a$ and for each fact $p \in Add(a)$, a rule $C{\rightarrow}p$ is formed, where $C=Pre(a)$. Assuming a set of rules, it is said that a fact $p$ is *reachable* from a state $S$ if $p \in S$ or there is a rule $C \rightarrow p$ such that each fact $q \in C$ is reachable from $S$.

So, a function $g(p,S)$ is defined, which inductively assigns a number $i$ to each fact $p$, where $i$ is an estimate of the number of steps needed to achieve $p$ from $S$, i.e. the distance of $p$ from $S$. More specifically, $g(p,S)$ is set to 0 for every fact $p \in S$, while $g(p,S)$ is set to $i+1$, $i \geq 0$, for each fact $p$ for which a rule $C \rightarrow p$ exists, such that $\sum_{r \in C} g(r,S) = i$. Thus:

$$g(p,S) \stackrel{def}{=} \begin{cases} 0, & \text{if } p \in S \\ i{+}1, & \text{if for some } C{\rightarrow}p,\ \sum_{r \in C} g(r,S) = i \\ \infty, & \text{if } p \text{ is not reachable from } S \end{cases} \tag{4}$$

In the case where there are more than one rules $C{\rightarrow}p$ for a fact $p$, the rule with the minimum cost is chosen. Note that a fact $p$ that was initially achieved by a rule $C_1{\rightarrow}p$, may be re-achieved, later, by another rule $C_2{\rightarrow}p$ with smaller cost. That is because not all the preconditions of the second rule had been achieved at the time when the first rule was applied. The task of applying rules continues until no rule that can achieve a fact with smaller cost exists. The distances computed in this way are unique.

For a set of facts $P$, their distance from $S$ is defined as:

$$g(P,S) \stackrel{def}{=} \sum_{p \in P} g(p,S) \tag{5}$$

The ASP planner uses $g(P,S)$ to estimate the distances between each intermediate state $S$ and the *Goals*. So, the ASP heuristic function is defined as:

$$H_{ASP}(S) \overset{def}{=} g(Goals, S) \qquad\qquad \textbf{(6)}$$

The ASP heuristic does not take into account the delete lists of the actions. The simplified problem that is created by ignoring the delete lists is referred to as the *relaxed problem* and the corresponding actions are referred to as *relaxed actions*. The complexity for constructing $H_{ASP}(S)$ is linear, with respect to the number of ground actions and the number of ground facts.

## 2.2 Backward Heuristic Construction

Instead of estimating the distance between each fact and the current state in a forward direction, as ASP does, GRT estimates the distance between each fact and the goals in a backward direction. This task is performed once, in a pre-processing phase. During the search phase, these estimates are used to estimate the distance between each intermediate state and the goals. The backward or forward estimation of the distance between two states often results in different values, since no heuristic is precise. However, the two directions result in estimates of equal quality on average.

The estimates of the distances between each fact and the goals are stored in a table, the records of which are indexed by the facts. We call this table the Greedy Regression Table (by which the acronym GRT comes from), since its estimates are obtained through greedy regression from the goals.

In order to construct the heuristic backwards, the actions of the problem have to be inverted. Let $a$ be an action and $S$ and $S'$ be two states, such that $a$ is applicable in $S$ and $S' = res(S,a)$. The *inverted action a'* of $a$ is an action applicable in $S'$, such that $S = res(S', a')$. The inverted action is defined by the original action as follows:

$$Pre(a')=Add(a) \cup Pre(a) \setminus Del(a)$$
$$Del(a')=Add(a) \qquad\qquad \textbf{(7)}$$
$$Add(a')=Del(a)$$

The inverted ground actions are applied to the goals, assigning progressively to each ground fact $p$ an estimate of its distance from the goals, in a way similar to ASP. Applying inverted actions to the goals presupposes that the goals form a complete state. In Section 2 it is assumed that this is always the case, whereas in Section 3 the case of incomplete goal states is treated.

## 2.3 Related Facts

In order to obtain more precise estimates, GRT heuristic tries to track the interactions that arise when estimating the distances between each fact and the goals. By the word 'interaction' we mean that achieving a fact may affect achieving other facts positively or negatively. In order to track these interactions the notion of the *related facts* is introduced.

**Definition 1 (Related facts).** A fact $q$ is related to another fact $p$, if achieving $p$ causes fact $q$ to be achieved as well.

We will use the notation $q \prec_{rel} p$ to denote that $q$ is related to $p$. The set of all facts related to a specific fact $p$ is denoted as $rel(p)$, i.e.:

$$rel(p) = \{q : q \prec_{rel} p\} \qquad\qquad \textbf{(8)}$$

The set of related facts of a set of facts $P$ is defined as the union of the related facts of $P$-facts:

$$rel(P) = \bigcup_{p \in P} rel(p) \qquad\qquad (9)$$

**Proposition 1.** For an inverted action $a$ achieving a fact $p$, the related facts of $p$ are defined as:

$$rel(p) = Pre(a) \cup rel(Pre(a)) \cup Add(a) \setminus Del(a) \qquad\qquad (10)$$

**Proof:** Formula 10 is inductive, since it defines the related facts of a fact $p$ based on the related facts of the preconditions of the action achieving the fact. Thus, we prove it by induction. The formula holds for the goal facts, for which we suppose that there is a hypothetical inverted action without preconditions achieving them. So, the goal facts are related to each other. Then, suppose that Formula 10 holds for the preconditions of an inverted action $a$. It is enough to prove that it holds also for the facts that action $a$ adds. Let $p$ be such a fact. The facts that hold after the application of the action, which are the related facts of $p$, are the same that hold before its application, i.e. the preconditions of the action together with their related facts, plus the facts that the action achieves, minus the facts that the action deletes, exactly as Formula 10 states. ■

According to Formula 10, facts achieved by the same action have the same related facts. Moreover, each fact is at least related to itself.

If there was a single path to achieve a specific fact, then its related facts would be defined in a unique way. However, this is a rare situation. Thus, there are many actions that achieve a fact, many paths that achieve the preconditions of these actions; therefore, there is an extremely large number of possible combinations. Storing, for each fact, the related facts for all the possible ways of achieving it, requires huge amounts of time and space. For efficiency reasons we decided to store only one set of related facts for each fact, the set that corresponds to the shortest path that achieves the fact, according to the heuristic.

**Proposition 2.** The relation $\prec_{rel}$ is reflexive, but it is neither symmetric, nor transitive.

**Proof:** The relation $\prec_{rel}$ is reflexive, since each fact is related to itself. The relation $\prec_{rel}$ is not symmetric, since for a fact $q$, which is pre-requisite to achieve $p$, $q \prec_{rel} p$ may hold (if the action achieving $p$ does not delete $q$) while $p \prec_{rel} q$ may not hold, since $q$ may have been achieved before $p$. Finally, the relation $\prec_{rel}$ is not transitive, since from the relations $q \prec_{rel} p$ and $p \prec_{rel} r$ we cannot conclude that $q \prec_{rel} r$ holds, since it is possible for the action achieving $r$ to delete $q$. ■

For a fact $p$, $dist(p)$ denotes its estimated distance from the goals. Next, we present some axioms concerning the distances of the facts.

**Axiom 1.** The cost of achieving a set of facts $\{p_1, p_2, ..., p_N\}$ simultaneously, cannot be lower than the maximum of their individual distances.

$$dist(\{p_1, p_2, ..., p_N\}) \geq \max_{i=1}^{N} (dist(p_i)) \qquad\qquad (11)$$

**Axiom 2.** If an inverted action $a$ achieves a fact $p$, the distance of $p$ is equal to the cost of simultaneously achieving $a$'s preconditions plus one.

$$dist(p) = dist(\{p_1, p_2, ...\}) + 1, \text{ where } p_i \in Pre(a) \qquad\qquad (12)$$

**Proposition 3.** If $q \prec_{rel} p$ is true for two facts $q$ and $p$, then $dist(q) \leq dist(p)$.

**Proof:** We will prove Proposition 3 by induction. Proposition 3 holds for the *Goals*, since all the goal facts have zero distances and are related to each other. Suppose now that Proposition 3 holds for the set of the currently achieved facts *Facts*. It suffices to prove that for an action $a$, such that $Pre(a) \subseteq Facts$, Proposition 3 holds for the set $Facts \cup Add(a)$.

Suppose that there is a fact $p \in Add(a)$ that has just been achieved, or re-achieved with smaller cost. If there is another fact $q \in Facts \cup Add(a)$, such that $q \prec_{rel} p$, then either $q$ has also just been achieved by $a$ and hence $dist(q) = dist(p)$, or $q$ is a precondition of $a$ and then, according to Axiom2, $dist(q) < dist(p)$, or finally $q$ is a related fact of an $a$'s precondition, say $q'$ and then $dist(q') < dist(p)$ (Axiom 2) and $dist(q) \leq dist(q')$ (Proposition 3 holds for *Facts*), so $dist(q) < dist(p)$.

Let us suppose now that there is another fact $q$, such that $p \prec_{rel} q$. If $q$ has been achieved by $a$, then $dist(p) = dist(q)$. If $q$ has not been achieved by $a$, then $q$ has previously been achieved by another action, so $q \in Facts$. In this case, $p$ would also have been previously achieved by another action, before being re-achieved by $a$, so also $p \in Facts$. Since Proposition 3 holds for *Facts*, $dist_{OLD}(p) \leq dist(q)$, where $dist_{OLD}(p)$ the previous distance of $p$. But the new distance of $p$ is smaller than its previous distance, $dist(p) < dist_{OLD}(p)$, so $dist(p) < dist(q)$. Therefore, Proposition 3 holds in every case. ∎

**Corollary 1.** If $q \prec_{rel} p$ and $p \prec_{rel} q$, then $dist(p) = dist(q)$.

**Corollary 2.** If $q \prec_{rel} p$ but not $p \prec_{rel} q$, then $q$ has been achieved before $p$.

The above two corollaries follow directly from Proposition 3. Concerning Corollary 2, the expression 'has been achieved before' means that in the pre-processing phase, when the distances from the goals are estimated progressively, $dist(q)$ has been computed before dist($p$). In case where a fact has been re-achieved with smaller distance, we consider the last time.

**Corollary 3.** For a sequence of facts $p_1$, $p_2$, ..., $p_N$, N>2, for which $p_i \prec_{rel} p_{i+1}$, i=1,2,...,N-1, hold, without $p_{i+1} \prec_{rel} p_i$ also holding, it is impossible to have $p_N \prec_{rel} p_1$.

Corollary 3 follows directly from Corollary's 2 time ordering relation.

**Proposition 4.** Facts related to each other have been achieved by the same action.

**Proof:** Let $p$ and $q$ be two facts related to each other, i.e. $q \prec_{rel} p$ and $p \prec_{rel} q$. Let $a_1$ be the action that achieves $p$ and $a_2$ the action that achieves $q$, so $p \in Add(a_1)$ and $q \in Add(a_2)$. We will prove that $a_1 \equiv a_2$. Suppose that $a_1 \neq a_2$. Since $q \prec_{rel} p$, $q$ may be an add effect of $a_1$, a precondition of $a_1$, or a related fact of an $a_1$'s precondition. However, according to Corollary 1, $dist(p) = dist(q)$. Thus, $q$ cannot be anything else than an add effect of $a_1$, because in other case dist($q$) < $dist(p)$ would hold. In the same way we can prove that $p \in Add(a_2)$. Thus, $\{p,q\} \subseteq Add(a_1) \cap Add(a_2)$. However, in this case, the first action applied when computing the distances would achieve both facts. So, the facts have been achieved by the same action. ∎

The related facts play a critical role when estimating the cost of achieving a set of facts simultaneously. GRT groups the related facts and sums the maximum individual cost of each group. For example, if $q \prec_{rel} p$, $p \prec_{rel} r$ and $q \prec_{rel} r$ hold for three facts $q$, $p$ and $r$, these three facts are grouped together and contribute to the total cost only with their maximum cost, which is $dist(r)$. However, if $q \prec_{rel} r$ does not hold (since the relation $\prec_{rel}$ is not transitive), then $p$ and $r$ are grouped together, while $q$ is not included in the same group. In this case, $q$ belongs to another group, which contributes separately to the total cost.

The aggregation process is performed by the function AGGREGATE, which is described below. The function takes a set of facts $\{p_1, p_2, ...., p_N\}$ as input, together with their distances $dist(p_i)$ and their lists of related facts $rel(p_i)$, and estimates the cost of achieving them simultaneously. The function is used both in the pre-processing phase, in order to estimate the application cost of the inverted actions, and in the search phase, in order to estimate the distance of each intermediate state from the goals.

---

**Function AGGREGATE**

Input:    A set of facts $\{p_1, p_2, ..., p_N\}$, their distances $dist(p_i)$ and their lists of related facts $rel(p_i)$.

Output:   An estimate of the cost of achieving the facts simultaneously.

1. Set $M_1 = \{p_1,\ p_2,\ ...,\ p_N\}$. Set $Cost = 0$.

2. While ($M_1 \neq \emptyset$) do:

   a) Let $M_2$ be the set of facts $p_i \in M_1$ that are not included in any list of related facts of another fact $p_j \in M_1$, without $p_j$ being also included in their list of related facts. More formally:

$$M_2 = \{\ p_i: p_i \in M_1,\ \forall\ p_j \in M_1,\ p_i \in rel(p_j) \Rightarrow p_j \in rel(p_i)\ \}$$

   b) Let $M_3$ be the set of those facts of $M_1$ that are not included in $M_2$, but are included in at least one of the lists of related facts of the elements of $M_2$.

$$M_3 = \{\ p_i: p_i \in M_1 \setminus M_2,\ \exists\ p_j \in M_2,\ p_i \in rel(p_j)\ \}$$

   c) Divide $M_2$ in disjoint groups of facts that are related to each other. For each group add the common cost of its facts to $Cost$.

   d) Set $M_1 = M_1 \setminus (M_2 \cup M_3)$.

3. Return $Cost$

---

The AGGREGATE function is illustrated with the *blocks-world* problem of Figure 1. Part of the Greedy Regression Table for this problem is shown in Table 1. For simplicity, for each fact $p$ we do not consider as related the facts that have zero distances (i.e. the *Goals*) and the fact $p$ itself. This simplification does not affect the estimated distances.

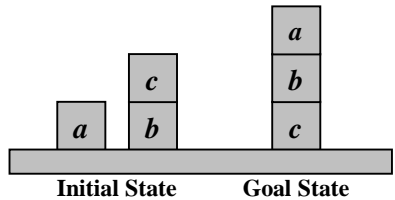

Figure 1: A 3-blocks problem.

Let us compute the distance between the initial and the goal state. The initial state consists of the following set of facts:

( (*on a table*) (*clear a*) (*on b table*) (*on c b*) (*clear c*) )[1]

As it results from Table 1, all the initial state facts are related to (*on c b*), whereas (*on c b*) is not related to any other fact. Thus, in the first iteration of the AGGREGATION loop, $M_2$ is set to ((*on c b*)) (step 2a) and $M_3$ is set to ((*on a table*) (*clear a*) (*on b table*) (*clear c*)) (step 2b). So, *Cost* becomes equal to the distance of (*on c b*), i.e. 3 (step 2c) and $M_1$ becomes empty. A second iteration is not performed and value 3, which is the actual distance between the initial and the goal state, is returned.

| Fact | Distance from goals | Related facts |
|------|---------------------|---------------|
| (*on c table*) | 0 | ( ) |
| (*on b c*) | 0 | ( ) |
| (*on a b*) | 0 | ( ) |
| (*clear a*) | 0 | ( ) |
| (*on a table*) | 1 | ( (*clear b*) ) |
| (*clear* B) | 1 | ( (*on a table*) ) |
| (*on b table*) | 2 | ( (*on a table*) (*clear a*) (*clear b*) (*clear c*) ) |
| (*clear c*) | 2 | ( (*on a table*) (*clear a*) (*clear b*) (*on b table*) ) |
| (*on c b*) | 3 | ( (*on a table*) (*clear a*) (*on b table*) (*clear c*) ) |
| ... | ... | ... |

Table 1: Part of the Greedy Regression Table for the 3-blocks problem.

Corollary 3 ensures that set $M_2$ (step 2a of function AGGREGATE) will never be empty. Proposition 4 ensures that $M_2$ can always be partitioned in groups of facts that have been achieved by the same action (step 2c). The number of iterations that function AGGREGATE performs is bounded by the initial size of $M_1$, however usually a single iteration is performed.

## 2.4 The Pre-Processing Algorithm

The estimation of the distance between each fact and the *Goals* and the computation of the lists of the related facts for each facts of a problem are performed through the following algorithm:

---

[1] For the representation of facts, actions and states we adopt the PDDL (Planning Domain Definition Language) syntax throughout this paper. A manual for the PDDL language can be found at the URL
http://www.cs.yale.edu/pub/mcdermott/software/pddl.tar.gz

---

**The Pre-Processing Algorithm**

**Input:**　　　The action and predicate definitions of a domain and the objects of a problem.

**Output:**　　　The distance estimate from the goals $dist(p)$ and the related facts $rel(p)$ for each ground fact $p$ of a problem.

1. Let `Actions` be the set of all inverted ground actions in the given problem. For each $\alpha \in$ `Actions`, set `dist(`$\alpha$`)`=+∞.

2. Let `Agenda` be a list of inverted actions. Set `Agenda`=∅.

3. Let `Facts` be the set of all problem's ground facts. For each $f \in$ `Facts` set `dist(f)`= +∞.

4. For each $f \in$ `Goals` set `dist(f)`=0 and `rel(f)`=`Goals`.

5. For each action $\alpha \in$ `Actions`, if AGGREGATE(`Pre(`$\alpha$`)`)<+∞, then set `dist(`$\alpha$`)`=AGGREGATE(`Pre(`$\alpha$`)`)+1 and add $\alpha$ at the end of the `Agenda`.

6. While `Agenda` ≠ ∅ do:

　　a) Extract the first action from the `Agenda`, say $\alpha$.

　　b) For every fact $f \in$ `Add(`$\alpha$`)`, if `dist(f)`>`dist(`$\alpha$`)`, then:

　　　　– `dist(f)`=`dist(`$\alpha$`)`

　　　　– `rel(f) = Pre(`$\alpha$`)` ∪ `rel(Pre(`$\alpha$`))` ∪ `Add(`$\alpha$`)`\`Del(`$\alpha$`)`

　　　　– For every action $b \in$ `Actions`, such that $f \in$ `Pre(b)`, if AGGREGATE(`Pre(b)`)+1<`dist(b)`, then `dist(b)`=AGGREGATE(`Pre(b)`)+1 and push action $b$ at the end of the `Agenda`.

---

The *Agenda* works on a FIFO basis. An action can be re-inserted in the *Agenda* if its cost becomes smaller. Thus, each fact can be achieved several times, each time with a smaller cost. The cost of applying the Pre-Processing Algorithm is polynomial in the number of problem ground facts and ground actions.

**Proposition 5.** The Pre-Processing Algorithm preserves Axiom 2.

**Proof:** In step 6b, the cost of applying an action is set to be equal to the cost of achieving simultaneously the preconditions of the action plus one. This cost is assigned to the add effects of the action, except if lower costs have already been assigned to them. Thus, Axiom 2 is preserved. ∎

**Proposition 6.** Function AGGREGATE preserves Axiom 1.

**Proof:** We will prove Proposition 6 by induction. Axiom 1 holds for the *Goals*, which have zero distances from themselves and are related to each other. Besides, Propositions 3 and 4 and Corollaries 1, 2 and 3 hold also for them. Suppose next that Axiom 1 and all the induced Propositions and Corollaries hold for the currently achieved facts *Facts*. It suffices to prove that for any action $a$, such that $Pre(a) \subseteq Facts$, Axiom 1 holds for the new set of achieved facts $Facts'=Facts \cup Add(a)$.

Consider a set of facts $P \subseteq Facts'$. We will prove that function AGGREGATE preserves Axiom 1, with regard to the randomly selected set $P$. Let $p$ be the fact with the maximum distance among the facts of $P$. According to the definition of AGGREGATE function, it suffices to prove that $p$ or another fact of equal distance is included in $M_2$.

If $p \in P \backslash Add(a)$, then for every other fact $q \in P \backslash Add(a)$, if $p \prec_{rel} q$, then $dist(q) \geq dist(p)$ (according to Proposition 3, which holds for *Facts*) and finally $dist(q)=dist(p)$, because $p$ has the maximum distance among the facts of $P$ (the same rationale can be used in the case where there is a sequence of facts $q_1, q_2, ..., q_N$, such that $p \prec_{rel} q_1$ and $q_i \prec_{rel} q_{i+1}$, i=1, 2, ..., N-1). If $q \in Add(a)$ and $p \prec_{rel} q$, $p$ would be a precondition of $a$, or a related fact of a precondition of $a$. However, in that case it would not be possible that $p \prec_{rel} q$, because the distance of $q$ would be greater than the cost of $p$ (according to Axiom 2, which holds for the preconditions of action $a$) and this is in contradiction with the hypothesis that $p$ has the maximum distance among the facts of $P$.

Let us consider the case where $p \in Add(a)$. If $p$ has just been firstly achieved, then the only facts $q$, for which $p \prec_{rel} q$ hold, are certainly the other just achieved or re-achieved add effects of action $a$, which have the same application cost. If $p$ has been re-achieved by $a$ with smaller cost, then it is impossible to hold $p \prec_{rel} q$ for another fact $q \in P \backslash Add(a)$. Actually, in this hypothetical case we would have $dist(q) \geq dist_{OLD}(p)$, since Proposition 3 holds for $q$ and the previous distance of $p$, and $dist_{OLD}(p) > dist_{NEW}(p)$, so $dist(q) > dist_{NEW}(p)$, which is in contradiction with the hypothesis that $p$ has the maximum distance among the facts of $P$. Therefore, in any case, $p$ or another fact of equal cost is included in $M_2$ and the cost of achieving simultaneously the facts of $P$ is equal to or higher than their maximum distance. ∎

We close this section by mentioning the two types of facts, the *static facts* and the *dynamic facts*, that can be found in a problem. The first type concerns the facts that are neither added nor deleted by any action, while the second concerns the rest of the facts. GRT classifies automatically the facts, by analyzing the action schemas of the domain. All the procedures presented in Section 2, i.e. the distance estimates and the related facts, concern only the dynamic facts.

## 3. Detecting and Enhancing Incomplete States

Backward heuristic construction induces a problem: In most of the problems the goals do not constitute a complete state description, so it is not possible to apply inverted actions to them. For example, in the commonly used *logistics* problems, where packages have to be moved between several locations via trucks and planes, the goals do not determine the final locations of the trucks and the planes. The source of the problem is that the GRT heuristic is constructed using a stricter than usual regression, i.e. it uses actions, the add effects and the non-deleted preconditions of which (i.e. the preconditions of the corresponding inverted actions) are included within the goals (in the usual regression, actions with at least one add effect within the goals are used). In this way GRT succeeds in obtaining more precise estimates and avoiding unreachable facts.

The solution adopted by GRT to confront the problem of incomplete goal states is to enhance the goals with new facts, which are not in contradiction to the existing ones. For example, since the goals of the 'logistics.a' problem (Veloso, 1992) do not determine the final locations of the two planes, it is supposed that each one of the planes could be at any of the three airports. So, the ground facts:

(*at plane1 pgh_air*) (*at plane1 bos_air*) (*at plane1 la_air*)
(*at plane2 pgh_air*) (*at plane2 bos_air*) (*at plane2 la_air*)

can be added to the new goal state, which is called henceforth the *enhanced goal state*.

It should be noted that the enhanced goal state is only used in the pre-processing phase, for the construction of the heuristic. During the search phase, attention is paid only to reach the original goals. In this way, completeness is never lost, even in the case where wrong facts have been selected to enhance the *Goals*. However, selecting wrong facts may significantly affect the efficiency of the heuristic function.

Two issues arise when trying to enhance the goals: The first one is how to detect the candidate new goal facts and the second one is which of them to use. Sections 3.1 and 3.2 examine these issues, while in Section 3.3 a similar technique is used for identifying and enriching poor domain representations.

## 3.1    Detecting Missing Goal Facts

Regarding the identification of the candidate facts to enhance the goals, there are two automatic approaches. The first one consists of a forward GRAPHPLAN-like (Blum & Furst, 1999) pre-preprocessing phase that computes all binary mutual exclusion relations (or simply "mutex" relations) among the facts of the problem. A number of optimizations of this approach are presented in (Refanidis & Vlahavas, 1999c), based primarily on the monotonic behavior of the mutual exclusion relations (Long & Fox, 1999; Smith & Weld, 1999) and secondly on the fact that it is not necessary to construct a complete planning graph, since it will not be used for extracting a plan. After the computation of the mutual exclusion relations, all the facts that are not mutually exclusive with any goal fact are considered candidates for the enhancement of the goals. Its advantage is that no extra information is needed, apart from the usual STRIPS domain representation. Moreover, mutual exclusion relations that are not easily recognized by a human expert can be detected in this way. Finally, this approach can be also exploited as a coarse-grained reachability analysis for the problem's facts. The disadvantages of this approach are that it is time consuming and that it does not detect mutual exclusion relations of higher order than two.

The second approach is to use domain specific knowledge in the form of axioms. For example, an axiom can state that a truck or a plane is always located at some place. So, if the goals do not determine where a truck is, we can deduce a set of candidate goal facts using this axiom. The advantage of this approach is that the time needed to deduce the candidate facts is negligible, in comparison with the time needed for the rest of the planning process. Moreover, more complicated relations than simple binary mutual exclusion ones can be encoded. The disadvantage is that extra labor is required in the domain encoding. However, several methods for automatic discovery of domain axioms have been proposed, e.g. the DISCOPLAN system (Gerevini & Schubert, 1998) and the work of Fox and Long on the automated inference of invariants (Fox & Long, 2000), and it is in our future plans to adopt such a method in GRT.

The GRT planner uses the first approach to detect the missing goal facts. Thus, an overhead in total solution time is imposed by the extra pre-processing work. The contribution of this work to the total problem solving time varies from less than 10% in domains like *blocks-world*, to more than 20% in domains like *logistics*. The ratio depends on the difficulty of the domain, i.e. how much time is consumed by the search phase. Logistics problems are easier than *blocks-world* problems, so in this domain the overhead is more severe. In the future, we intend to adopt an automatic method for detecting domain axioms, in order to avoid this overhead.

## 3.2    Enhancing the Goals

GRT supports three methods of selecting among the candidate new goal facts:

- ▪  Select all candidate facts.
- ▪  Use the initial state facts.
- ▪  Favor the most promising facts.

The first method considers all the found facts as goal facts and assigns zero distances to them. In most cases, the enhanced goal state obtained in this way is not a valid state, since the new facts may be mutually exclusive to each other (but not to the original goals). The advantage of this approach is that the heuristic construction is very fast, since many facts are achieved at the beginning and a large number of actions become initially applicable. The disadvantage is that the obtained heuristic is less informative, since there are small differences between the obtained estimates. So, the best-first strategy tends towards breadth-first, visits more states, consumes more time, but generally produces better plans than the other two methods.

The second method enhances the goals with the candidate facts that are also included in the initial state, whereas the facts that are mutually exclusive with the selected ones, are rejected. The advantage of this method, compared to the first one, is that it results in greater differences between the facts' distances, and therefore in faster search phase. On the other hand, a preference for the initial state facts is a risk, because if these are not or - even worse - they cannot be included within the goals, the search process may become disoriented, leading to longer plans. This method is more suitable to problems, where there are objects' properties that are unnecessary to solve the problem and are left undetermined in the goals.

The third method tries to combine the advantages of the other two. In contrast to them, where the enhancement of the goals is performed in a single step, prior to the construction of the heuristic, this method adds facts to the goals progressively, in parallel with the heuristic construction. Actually, facts are added to the goals only in the case where *Agenda* (Section 2.4) becomes empty. In this case, candidate facts are progressively assigned zero distances, until a new inverted action satisfies its preconditions. Each time a fact is selected, other candidate facts that are mutually exclusive with the selected one are rejected from the set of candidate facts.

The method favors facts that can be combined with already achieved facts, in order to make an inverted action applicable. The following four rules are applied in decreasing preference:

- The facts that can be combined with the original goals are selected first.
- Then, the facts that can be combined with other already achieved facts are selected.
- Next, the facts that are included in the initial state are selected.
- Finally, the remaining candidate facts are selected randomly.

Generally, this method results in the best solving speed and, in many cases, produces equal or even better plans than the first two methods. However, especially in terms of plan quality, there are many exceptions depending on the specific problem. It is not difficult to create problems such that any of the methods presented above performs best. The default method for the GRT planner is the first one, which is the only method that has been used in the AIPS-00 competition[2].

Note that there are domains, like *blocks-world*, *freecell* and *elevator* of the AIPS-00 competition, or the *gripper* and the *movie* domains from the AIPS-98 competition[3], where the goals are complete or near-complete state descriptions; therefore the method used in these domains does not affect neither solution time nor solution quality. In other domains, as the *mystery* (AIPS-98), it is impossible to predict, without solving the planning problem, which of the candidate facts could actually be goal facts, so in this case the only acceptable method for goal completion is the first one.

---

[2] The official WEB page of the AIPS-00 competition can be found at the URL http://www.cs.toronto.edu/aips2000/.
[3] The official WEB page of the AIPS-98 competition can be found at the URL
   ftp://ftp.cs.yale.edu/pub/mcdermott/aipscomp-results.html

### 3.3    Domain Enrichment

In this section, we present an approach adopted by the GRT planner, in order to deal with poor domain descriptions. By the word 'poor' we refer to domains where negative facts are implicitly present in the initial state and in the actions' preconditions. GRT faced this problem twice, with the *movie* and the *elevator* domains.

In order to explain the problem, let us consider the *elevator* domain, where there is one elevator, several floors and several passengers. Each passenger is located in an initial floor and wants to move to her/his destination floor. The domain is described by four action schemas, (*board Floor Passenger*) and (*depart Floor* P*assenger*) for boarding and leaving the elevator and (*up Floor1 Floor2*) and (*down Floor1 Floor2*) for moving the elevator.

The action schema (*board Floor Passenger*) is defined by the following PDDL formula:

```
(:action board
:parameters (?f ?p)
:precondition (and (floor ?f) (passenger ?p)
 (lift-at ?f) (origin ?p ?f))
:effect (boarded ?p))
```

The only dynamic predicate in the definition of action schema *board* is *boarded*, an add effect denoting that the passenger has boarded the elevator. There is no precondition requiring that the passenger is not boarded. The problem with this definition is twofold. Firstly, the action can be applied several times to the same passenger in the same plan, i.e. a passenger may board the elevator although she/he has already boarded. Secondly, and specifically to GRT, it is not stated explicitly that the passengers are not initially boarded. Actually, the initial state contains static facts only, which are not removed in the successive states. However, GRT takes into account dynamic facts only in order to estimate distances. The result is that the initial state and all the subsequent states are assigned zero distances from the *Goals* and the best-first strategy behaves as a breadth-first one.

What is needed is the definition of a new predicate, say *not_boarded*. Facts of this predicate should be added to the initial state, denoting that each passenger is initially not boarded, and the action schema *board* should be changed accordingly.

GRT performs domain enrichment at run-time. The identification of the above situation is performed in a way similar to the identification of the incomplete goal states. In this case, GRT looks for dynamic facts of a problem that are not mutually exclusive with any initial state fact. In case of such facts, the negations of the identified facts are defined at run-time and added to the initial state. Furthermore, the negations are added to the preconditions lists and the delete lists of the actions that achieve the identified facts.

In the *elevator* domain this is the case with the *board* and *depart* actions and the *boarded* and *served* predicates. The *not_boarded* and *not_served* predicates are defined at run-time, the initial state is enhanced with facts determining that each passenger is neither boarded nor served yet and the actions *board* and *depart* are transformed accordingly. For example, the action schema *board* is transformed into the following definition:

```
(:action board
:parameters (?f ?p)
:precondition (and (floor ?f) (passenger ?p)(lift-at ?f)
 (origin ?p ?f) (not_boarded ?p))
:effect (and (not (not_boarded ?p))(boarded ?p))
```

A similar situation arises in the *movie* domain. In this domain, the goal is to have enough snacks so as to watch a movie. There are several action schemas of the form:

```
(:action get-chips
        :parameters (?x)
        :precondition (and (chips ?x))
        :effect (and (have-chips)))
```

This action schema has the static fact (*chips ?x*) as precondition and produces the dynamic fact (*have-chips*). The action can be applied several times, however once is enough to achieve the goal of having chips. The difficulty in this domain is that the initial state implicitly declares that we do not have chips (and dips and pops etc), but there is not any specific dynamic fact to make this clear. Therefore, in case no domain enrichment process takes place, GRT assigns to the initial state a zero distance from the goals. With the domain enrichment feature, GRT detects that there are facts like the *have-chips*, *have-dips* etc that are not mutually exclusive with the initial state, defines their negations (*not_have-chips*, *not_have-dips* etc.), adds them to the initial state and transforms the actions accordingly.

In both of the above domains, without the domain enrichment feature the GRT planner could only solve some of the easiest problems. However, with this feature it was able to tackle all problems very efficiently.

Adding negative predicates in the preconditions of the actions may lead to loss of completeness, since the actions may not be able to be applied in some states, where otherwise they could. In order to prevent completeness, GRT treats the new preconditions as conditional preconditions, i.e. they are not necessary for the application of an action to a state, however, if they are present in the current state they are removed from the successor one.

## 4. Reducing the Size of the Problems

In this section, two methods to reduce the size of a problem, i.e. the number of ground facts and actions, are presented. The first method refers to the identification and elimination of objects, which are certainly not part of any solution. The second method concerns the adoption of a numerical representation of resources, instead of the problematic atom-based representation of numbers that has been used in domains like *mystery* and *freecell*. Reducing the size of a problem reduces the effort needed to solve it, especially in the pre-processing phase, where distances for all facts of a problem have to be computed.

### 4.1 Eliminating Irrelevant Objects

In many domains, there are objects that are irrelevant to any solution. The most typical examples can be found in the transportation domains, like *logistics*, *mystery* and *elevator*, where some packages are initially found in their destinations or for which no specific destination is determined. So, these objects, together with all the facts and actions containing them, can be removed from the problem description, without losing completeness.

In GRT we developed a method that detects and removes irrelevant objects. The method concerns pure STRIPS domains without negation in the preconditions of the actions or in the goal formula; however, it can be easily extended to cover these cases. The objects are identified before the pre-processing phase using the following two rules:

An object is irrelevant to any solution for a specific planning problem, if:

- It does not appear in any goal fact, unless the same fact is also included in the initial state, and
- there is no action containing this object in its preconditions, unless the object is also contained in all the action's effects.

The above conditions are very strict, but they ensure that any detected object is certainly irrelevant, so they maintain the completeness of the problem solving process.

**Proposition 7.** Any object satisfying the above rules can safely be removed from the problem description, without sacrificing completeness.

**Proof:** Suppose that an object *obj* has been identified, for which the above two rules hold. We will show that *obj* is not necessary to achieve any other goal fact, which does not contain *obj*. Let us assume that there is a fact *g* ∈ *Goals*, which does not contain *obj*. Suppose that there is an action that achieves *g*, with a precondition containing *obj*. In this case, the second rule is violated, since there is an action including *obj* in its preconditions, without *obj* appearing in an effect. So, fact *g* can be achieved only by actions without preconditions containing *obj*. Thus, if we regress the goals using actions achieving *g*, the established subgoals do not contain *obj*. However, in the same way we can reject actions including *obj* in their preconditions and achieve the new established subgoals. So, *obj* is not necessary to achieve any goal or subgoal of the problem. Moreover, there is no goal fact containing *obj*, which has to be achieved; even if there is one, it is already present in the initial state. Therefore, *obj* can safely be removed from the problem. ∎

The application of the above rules for the elimination of irrelevant objects can be done progressively. Let us consider an enhanced *logistics* domain, where we added colors. Specifically, we define a dynamic predicate (*painted ?object ?color*) denoting the color of a package, a static predicate (*color ?color*) declaring the available colors, and an action schema (*paint ?object ?old_color ?new_color*) for painting a package. Let us assume that the goal state does not determine the colors of the packages. In this case, the colors are irrelevant objects and can be safely removed, together with all the facts and actions that include colors.

Suppose also that there are brushes that are used to perform the paint operation. There are two new action schemas, (*get ?brush*) and (*leave ?brush*) and a predicate (*have ?brush*), which is an effect of the *get* action and a precondition in the enhanced action (*paint ?package ?color ?brush*). In this case, brushes are also irrelevant and should be eliminated. However, since the action *paint* needs brushes and has effects not containing them (i.e. (*painted ?package ?color*) ), the brushes are not removed, due to the second rule. However, after removing all the *color* objects, all the paint actions are removed; thus, brushes do not violate the second rule for the remaining actions and can be safely removed.

The disadvantage of this approach for the elimination of irrelevant objects is that it does not remove objects that can eventually appear in a plan, but there are other better (i.e. shorter) plans not using them. For example, in the *logistics* domain, suppose that we have three cities, *city1*, *city2* and *city3* and a package that has to be transferred from one location of *city1* to another location of *city2*. In this case, *city3*, together with its locations and its truck, are not necessary to solve the planning problem, since the package can be transferred directly from *city1* to *city2*, without going via *city3*. However, it is not easy to identify the irrelevance of *city3*. Actually, there are plans that transport packages from *city1* to *city2* via *city3*. If we decide to remove *city3* and its objects from the problem representation, we take the risk of sacrificing completeness, since the problem may become unsolvable. Deciding safely, without loss of completeness, that *city3* and its objects can be removed, can be as hard as solving the original problem.

Other approaches on the elimination of irrelevant or redundant information, in order to achieve better performance, have been proposed by Nebel, Dimopoulos & Koehler (1997), Scholz (1999) and Haslum & Jonsson (2000). The work of Nebel, Dimopoulos & Koehler concerns ignoring irrelevant facts and actions (not objects), based on heuristics that approximate a plan by backchaining from the goals without taking into account any conflicts. Although this approach is more powerful, in terms of elimination, than the one presented in this section, it is not solution preserving. Furthermore, it may be more time-consuming, since it demands the construction of an initial approximate plan.

Scholz introduces *action constraints*, i.e. patterns of action sequences that can be applied to the same states and produce the same overall effects. Action constraints can be used for pruning purposes by the state-space planners, reducing the size of the search space to the levels of the partial-order planners (Minton, Bresina & Drummond, 1994), without losing completeness. The work of Scholz is actually a re-investigation of the *sleep sets* of actions that were originally presented by Godefroid & Kabanza (1991) and have been also examined by us, under the name *prohibited actions*, in an earlier version of GRT (1999a). The experience of the authors is that detecting and pruning redundant actions sequences is time consuming, while a more effective approach is to employ a closed list of visited states, paying however a cost in terms of memory. The latter approach is adopted by the GRT planning system. However Scholz considers only action sequences of length two, which makes his approach fast enough but less effective than a closed list of visited states structure.

Haslum and Jonsson compute a reduced set of actions for a problem, by ignoring actions that can be equivalently replaced by sequences of other actions. Their approach is solution preserving, it can be adopted by any STRIPS planner that pre-instantiates all the actions of a problem, and results, for some planners, in considerable speed-up but also in longer plans.

## 4.2    Numerical Representation of Resources

In this section, we present an enhanced STRIPS formalism, where resources are represented by numbers, instead of atoms. The work has been motivated by the *mystery* domain, but it is suitable for any domain with resources. Moreover, it can easily be extended to cover domains where reasoning with numbers is required.

GRT supports an explicit representation of resources in the most natural format, i.e. the numerical format. According to this, resources are distinguished from other types of objects and are separately declared using the following statement:

    (:resources R1 R2 ... RN )

where *Ri* are the various resources. Furthermore, declarations of the following form are added to the initial state description :

    (amount R1 V1) (amount R2 V2) ... (amount RN VN)

denoting the initial quantity of each resource. Moreover, it is allowed for resources to participate in relations with other atomic facts. Finally, action definitions are enhanced, so as to declare explicitly the consumed resources.

As an example, we consider the *mystery* domain, which comprises some cities, connected via edges, some packages that have to be transferred from their initial locations to their destinations and some trucks. In the beginning, each city has an amount of fuel. For a truck to travel from a city *c1* to an adjacent city *c2*, *c1* must have at least one unit of fuel. After the journey, the fuel of *c1* is decreased by one.

In the original domain representation, the different fuel quantities are represented by relations of the form[4]:

> (`fuel fuel0`) (`fuel fuel1`) (`fuel fuel2`) etc.

while the orderings between these quantities are represented by relations as follows:

> (`adjacent_fuel fuel0 fuel1`) (`adjacent_fuel fuel1 fuel2`) etc.

and the initial amount of resources in each city as:

> (`city_fuel city1 fuel3`) etc.

Finally, the actions that consume resources, e.g. moving a truck, are of the following form:

```
(:action move
:parameters (?tr ?c1 ?c2 ?f1 ?f2)
:precondition (and (truck ?tr) (city ?c1) (city ?c2)
(adjacent_cities ?c1 ?c2) (fuel ?f1) (fuel ?f2) (at ?tr ?c1)
(adjacent_fuel ?f1 ?f2) (city_fuels ?c1 ?f2))
:effect (and (not (at ?tr ?c1)) (not (city_fuel ?c1 ?f2))
(at ?tr ?c2) (city_fuel ?c1 ?f1)))
```

In order to have an idea of how resources are represented in GRT, let us consider the STRIPS-MYSTY-X-1 problem of the *mystery* domain. This problem has 6 cities, so 6 resource objects are declared:

> (`:resources r1 r2 r3 r4 r5 r6`)

The resources are related with their corresponding cities:

(`city_fuel city1 r1`) (`city_fuel city2 r1`) ... (`city_fuel city6 r6`)

Propositions are added to the initial state, denoting the initial availability of each resource:

> (`amount r1 1`) (`amount r2 2`) ... (`amount r6 3`)

Finally, action *move* is defined in a way that separates the resource requirements from the precondition and the effect lists:

```
(:action move
:parameters (?tr ?c1 ?c2 ?f)
:precondition (and (truck ?tr) (city ?c1) (city ?c2) (at ?tr ?c1)
(adjacent_cities ?c1 ?c2) (city_fuel ?c1 ?f))
:effect (and (not (at ?tr ?c1)) (at ?tr ?c2))
:resources (amount ?f 1))
```

Table 2 shows the number of ground facts and ground actions for the first five problems of the *mystery* distribution, for the two alternative resource representations. As it is clear from this table, through the numerical representation of resources there is an important reduction in the number of ground facts, which is more considerable in the case of ground actions. What is even more important is that the size of the problem in the atom-based representation can grow illimitably, if more levels of resource availability are added, whereas in the numerical representation the size of the problem remains constant.

---

[4] In the AIPS-98 competition, different predicate and object names have been used; however, in this paper we have translated them into more meaningful ones for simplicity.

| | Atom representation | | Numerical Representation | |
|---|---|---|---|---|
| Problem | ground facts | ground actions | ground facts | ground actions |
| strips-mysty-x-1 | 101 | 150 | 56 | 48 |
| strips-mysty-x-2 | 359 | 3596 | 310 | 1200 |
| strips-mysty-x-3 | 277 | 1676 | 230 | 816 |
| strips-mysty-x-4 | 178 | 210 | 144 | 168 |
| strips-mysty-x-5 | 299 | 2325 | 269 | 1032 |

Table 2: Size of the problem (number of ground facts and actions)
for the two alternative resource representations.

## 5. Using XOR Constraints to avoid Local Optimal States

In this section, we tackle the problem of local optimal states. Firstly, we illustrate the problem, then we introduce XOR-constraints and finally we present how these are exploited by GRT in order to avoid local optima.

### 5.1 Local Optimal States

During the search phase, GRT always selects to expand the most promising state, according to its heuristic. If the various facts of a problem were independent or even if GRT always managed to track their interactions through the related facts, this strategy would be optimal. However, this is not always the case and some times the search is led to local optimal states. Therefore, the planner should temporarily backtrack to less promising states, before selecting the most promising ones. Figure 2 presents an example situation:

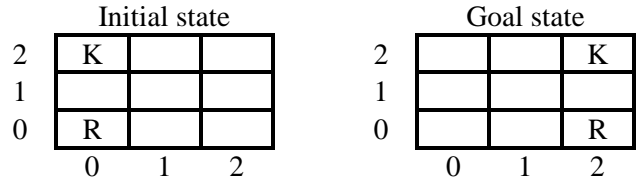

Figure 2: A 3x3 *grid* problem.

The problem refers to a *grid*-like domain (McDermott, 1999), where *K* is a key and *R* is a robot. The robot can only proceed to adjacent positions. The valid actions are *get* and *leave* the key and *move* the robot. Table 3 shows part of the Greedy Regression Table for the problem of Figure 2.

According to this Table, the distance between the initial and the goal state is 10. There are two applicable to the initial state actions, moving *R* to *n1_0* and moving *R* to *n0_1*. After moving *R* to *n1_0* the resulting state has a distance from the goals equal to 9, whereas after moving *R* to *n0_1* the resulting state has a distance from the goals equal to 11. So the planner decides to move *R* to *n1_0* and subsequently to *n2_0*. However, it is obvious that the optimal first movements are moving the robot to *n0_1*, next to *n0_2*, getting the key etc.

| Fact | Distance from Goals | Related Facts |
|------|---------------------|---------------|
| (*at R n2_0*) | 0 | ( ) |
| (*at K n2_2*) | 0 | ( ) |
| (*at R n1_0*) | 1 | ( ) |
| (*at R n0_0*) | 2 | ( ) |
| (*at R n0_1*) | 3 | ( ) |
| (*at R n2_1*) | 1 | ( ) |
| (*at R n2_2*) | 2 | ( ) |
| (*in R K*) | 3 | ( (*at R n2_2*) ) |
| (*at R n1_2*) | 3 | ( ) |
| (*at K n1_2*) | 7 | ( (*at R n1_2*) ) |
| (*at R n0_2*) | 4 | ( ) |
| (*at K n0_2*) | 8 | ( (*at R n0_2*) ) |

Table 3: Part of the Greedy Regression Table for the 3x3 grid problem.

Initially the planner does not select the optimal action, since it leads to a state with a greater distance from the goals, according to the heuristic. In order to decide to move the robot towards the key, the planner should go through all the other valid plans, then backtrack and move the robot to worse states (this requires that the planner maintains a closed list of visited states and does not revisit them). In difficult problems, the number of states that the planner has to visit before following the optimal direction, is extremely large. This is the main reason why GRT, like many other heuristic planners, does not handle *grid*-like domains efficiently.

For the 3x3 *grid* problem of Figure 2, an ideal planner should detect that, in order to move the key from *n0_2* to *n2_2*, it is necessary that the robot gets the key, so the fact (*at R n0_2*) should be achieved before the fact (*at R n2_0*). However, the planner does not know that the facts (*at R n0_0*), (*at R n2_0*) and (*at R n0_2*) are related in some way, because the domain definition does not provide this piece of information. Therefore, it is necessary to provide the planner with information about relations that hold between the facts of the problem.

## 5.2 Defining XOR-constraints

In order to avoid local optimal states, we provide GRT with knowledge of relations between facts, where exactly one of the facts can hold in each state. We call these relations XOR-constraints.

**Definition 2 (XOR-constraint).** An XOR-constraint is a relation between ground facts. The relation is valid in a state, if exactly one of the participating facts holds in that state.

The general form of an XOR-constraint schema is the following:

$$((xor\ f_1\ f_2\ ...)\ c_1\ c_2\ ...)$$

where $f_i$ are the facts that cannot co-appear in any state and $c_i$ are static facts that provide supplementary conditions such as type constraints, relations between objects, etc.

XOR-constraints can be formalized for almost any domain. For example, in the *logistics* domain we could define the following XOR-constraints:

( (*xor* ( *at ?Truck* * ) ) ( *truck ?Truck* ) )
( (*xor* ( *at ?Plane* * ) ) ( *plane ?Plane* ) )
( (*xor* ( *at ?Package* * ) ( *in ?Package* * ) ) ( *package ?Package* ) )

Question marks (?) precede named variables, whereas asterisks (*) denote no-named ones. The definitions mean that for every instantiation of the named variables that appear in an XOR-constraint and for all the valid instantiations of the no-named variables, according to the predicate definitions, exactly one ground fact can hold in each valid and complete state. The above XOR-constraints schemas are general definitions that can be grounded in several ways, according to the different ways in which their named variables can be instantiated.

In some cases, it is possible to have XOR-constraints that incorporate AND relations. For example, if in the *logistics* domain the predicate (*out ?Package*) is defined, which means that a package is not loaded either in a truck or in a plane, then the relevant constraint should be written:

( ( *xor* ( *and* ( ( *at ?Package* * ) ( *out ?Package* ) ) ) ( *in ?Package* * ) ) ( *package ?Package* ) )

Note that some facts may not appear in any XOR-constraint, while some others may appear in more than one. Henceforth, we refer to facts that appear in at least one XOR-constraint as *XOR-constrained facts*.

It is a requirement of the current version of GRT that the XOR-constraints are included in the domain definition. However, they could be computed analytically, based on the mutual exclusion relations between the facts of a problem, since mutually exclusive facts cannot appear simultaneously in any valid state. However, providing them manually allows for some form of guidance, since the domain engineer can leave out some of them, since they would lead to pointless decompositions.

The notion of XOR-constraints is not new in planning. Gerevini and Schubert (1998) proposed a method for the automatic inference of state constraints from the action definitions and the initial state. *Single valuedness constraints* or *sv constraints* are the closest to the XOR-constraints. But *sv constraints* concern instantiations of the same predicate, while XOR-constraints can be relations between ground facts of different predicates. However, in more recent work (2000a, 2000b), they extended their work to also infer XOR-constraints.

The object oriented domain specification formalism introduced by McCluskey & Porteous (1997) is similar to XOR-constraints. According to this, states are not defined as collections of facts but as collections of objects, each object having its own internal status. So, XOR-constraints can be implicitly defined from the requirement that all object attributes are single valued.

### 5.3 Decomposing Problems into Sub-problems using XOR-constraints

In this section we illustrate how GRT exploits XOR-constraints within the pre-processing phase, in order to avoid local optimal states. Specifically, using them GRT manages to establish new ordered subgoals that have to be achieved before achieving the original goals. These subgoals are grouped into ordered intermediate states, thus the original difficult problem is decomposed in a sequence of easier subproblems that have to be solved sequentially.

We will present the steps of the problem decomposition process through the example of Figure 3, a 4x4 *grid* problem with two keys (*K1* and *K2*) and two robots (*R1* and *R2*).

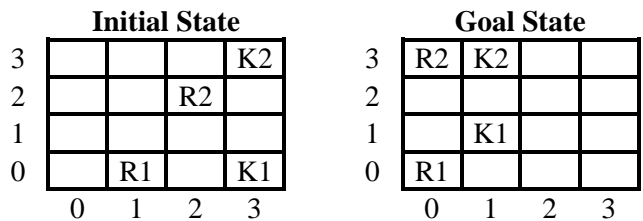

Figure 3: A 4x4 *grid* problem.

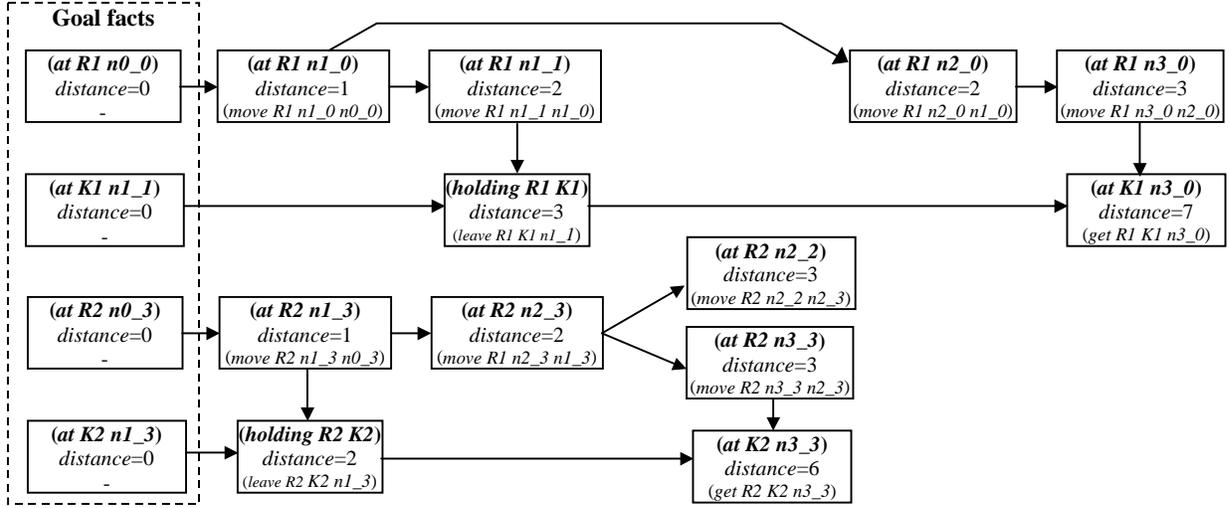

Figure 4: Part of the Greedy Regression Graph for the 4x4 Grid problem.

For this domain the following XOR-constraints can be defined:

   ( ( *xor* ( *at ?Robot* * ) ) ( *robot ?Robot* ) )
   ( ( *xor* ( *at ?Key* * ) ( *holding ?Key* ) ) ( *key ?Key* ) )

The above definitions have four ground instantiations, one for each *Robot* and one for each *Key*. Henceforth the notation XOR$_{OBJ}$ will refer to the ground XOR-constraint concerning object OBJ.

The first information that can be extracted is pairs of facts, one from the initial state and one from the goals, which belong to the same ground XOR-constraint. For the problem of Figure 3 the following pairs can be identified:

   **XOR$_{R1}$:** (*at R1 n1_0*) - (*at R1 n0_0*)
   **XOR$_{R2}$:** (*at R2 n2_2*) - (*at R2 n0_3*)
   **XOR$_{K1}$:** (*at K1 n3_0*) - (*at K1 n1_1*)
   **XOR$_{K2}$:** (*at K2 n3_3*) - (*at K2 n1_3*)

The original GRT planner did not store information about the inverted actions, which achieved the various facts in the heuristic construction phase. However, in order to exploit the XOR-constraints, this information has to be stored. By storing these actions, the table structure used by the GRT heuristic is transformed to a directed acyclic graph. We call this structure Greedy Regression Graph or simply GRG.

The nodes of this graph are labeled with the facts of the problem. Each node retains also the estimated distance between its fact and the goals and the corresponding related facts. It retains also the name of the inverted action that achieved its fact. The arcs that point to a node originate from the nodes of the preconditions of the inverted action that achieved the node's fact. Figure 4 shows part of the GRG structure for the 4x4 *grid* problem (the related facts are omitted).

Based on GRG, for every ground XOR-constraint, a sequence of actions which is able to transform the initial state fact to the corresponding goal state fact can be derived. We are interested only in the actions that change the XOR-constraint's facts and not in actions that provide auxiliary preconditions. For the problem of Figure 3, the actions' sequences are shown in Table 4:

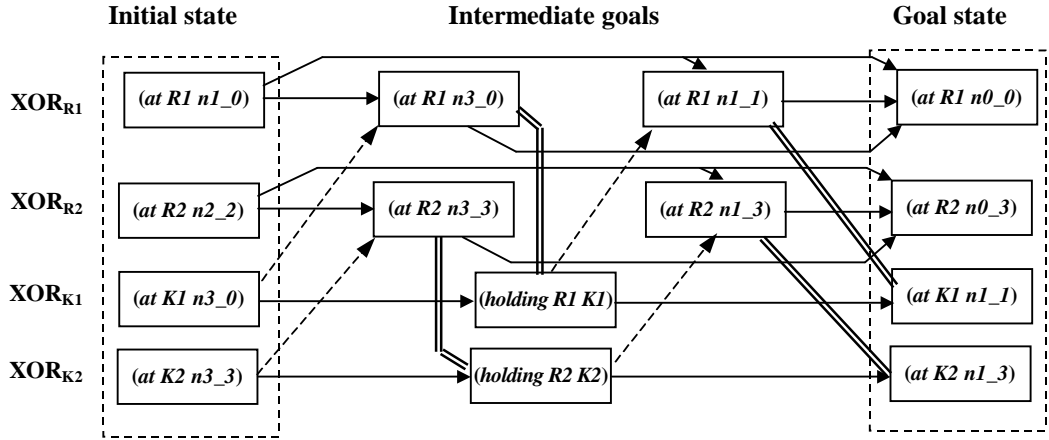

Figure 5: The *ordering graph* for the 4x4 grid problem.

| XOR constraints | Initial State Facts | Goal State Facts | Sequences of actions |
|---|---|---|---|
| XOR$_{R1}$ | (*at R1 n1_0*) | (*at R1 n0_0*) | (*move R1 n1_0 n0_0*) |
| XOR$_{R2}$ | (*at R2 n2_2*) | (*at R2 n0_3*) | (*move R2 n2_2 n2_3*) (*move R2 n2_3 n1_3*) |
| | | | (*move R2 n1_3 n0_3*) |
| XOR$_{K1}$ | (*at K1 n3_0*) | (*at K1 n1_1*) | (*get R1 K1 n3_0*) (*leave R1 K1 n1_1*) |
| XOR$_{K1}$ | (*at K2 n3_3*) | (*at K2 n1_3*) | (*get R2 K2 n3_3*) (*leave R2 K2 n1_3*) |

Table 4: Sequences of actions that transform the initial state facts
to the corresponding goal facts.

Checking the preconditions of the above actions, we can find facts that are members of foreign XOR-constraints. These facts are subgoals that have to be temporarily established, before achieving the original goals, in the forward search phase. In Table 4, the actions (*get R1 K1 n3_0*) and (*leave R1 K1 n1_1*) of the XOR$_{K1}$ sequence have (*at R1 n3_0*) and (*at R1 n1_1*) as preconditions respectively, which are members of the XOR$_{R1}$ relation. Similarly, the actions (*get R2 K2 n3_3*) and (*leave R2 K2 n1_3*) of the XOR$_{K2}$ sequence have (*at R2 n3_3*) and (*at R2 n1_3*) as preconditions respectively, which are members of the XOR$_{R2}$ relation.

There are two types of subgoals. These are the XOR-constrained facts that are either:

(I)     preconditions of a ground action in a foreign XOR sequence, or
(II)    add-effects of an action, in their own XOR sequence, which has a foreign precondition.

From the identified subgoals, we can construct a graph, conjoining the new subgoals with arcs that denote ordering constraints, using the following rules:

1.  All the subgoals are ordered after their initial state fact and before their goal fact (if any).
2.  Subgoals of type (II) that are members of the same XOR-constraint are ordered according to the ordering of their actions.
3.  Subgoals of type (I) are ordered together with the corresponding subgoals of type (II), which have resulted by the same action.
4.  For a specific XOR-constraint, subgoals of type (I) are ordered before the subgoals of type (II).

We call the resulted graph the *ordering graph* of the problem, since it denotes the order in which the subgoals have to be achieved. Figure 5 shows the *ordering graph* for the problem of Figure 3. Lines with arcs denote ordering constraints. Double-lines without arcs denote that the two facts are ordered together.

**Proposition 8**. The ordering graph is an acyclic graph.

**Proof sketch:** The proof can be based on the way in which the facts are achieved in the Pre-Processing Algorithm (Section 2.4). Actually, facts are achieved in a specific time order (in case where a fact has been re-achieved with smaller cost, we consider the last time it has been achieved). We define the ordering relation < between facts, denoting that a fact has been achieved before another in the Pre-Processing Algorithm. Similarly we define the ≤ relation.

Ordering relations between the subgoals originate in two ways. Firstly, subgoals of type (II) of the same XOR-constraint are ordered explicitly to each other, according to the time they have been achieved (in Figure 5 these ordering relations are denoted with non-dashed lines with arcs). Secondly, each subgoal of type (I) is ordered before than or at least at the same time with the previous one of its corresponding type (II) subgoal (in Figure 5 these ordering relations are denoted with dashed lines with arcs). Using the above equivalences, we can transform the ordering graph to an equivalent time-ordering graph. Since a time-ordering relation cannot include cycles, the same happens for the ordering graph. ∎

The ordering graph makes it possible to construct intermediate, possibly incomplete, states, which have to be achieved sequentially. Starting from the initial state, GRT attempts to insert one subgoal from each XOR-constraint in each intermediate state. This fact must have the following properties:

- It has not been inserted in a previous intermediate state,
- it is not ordered after some other fact of the same XOR-constraint that has not yet been inserted in a previous intermediate state, and finally
- it is not ordered together with a fact of another XOR-constraint that cannot be inserted in the current intermediate state.

In case where there are more than one facts with the above properties for a single XOR-constraint, the selection among them is done arbitrarily. Finally, in case where no fact with the above properties exists for an XOR-constraint, the intermediate state is left incomplete.

**Corollary 4.** It is always possible to construct the intermediate states.

Corollary 4 follows from Proposition 8. Since the ordering graph is a directed acyclic graph, it is always possible to find at least one subgoal to be included in the next intermediate state. The number of subgoals is an upper bound for the number of the intermediate states that will be constructed.

From the ordering graph of Figure 5, the following intermediate states can be extracted:

**Intermediate state 1:** ( (*at R1 n3_0*) (*at R2 n3_3*) (*in K1 R1*) (*in K2 R2*) )

**Intermediate state 2:** ( (*at R1 n1_1*) (*at R2 n1_3*) (*at K1 n1_1*) (*at K2 n1_3*) )

**Intermediate state 3:** ( (*at R1 n0_0*) (*at R2 n0_3*) (*at K1n1_1*) (*at K2 n1_3*) )

where the last state is the goal state.

After the construction of the intermediate states, the planner has to solve three sub-problems, which are easier than the original one; thus, the overall time to solve them is shorter than the time needed to solve the original problem. Note, however, that this decomposition may lead to loss of completeness. In domains where no deadlock exists, some solutions may be pruned. In domains where deadlocks do exist, the decomposition may produce unsolvable sub-problems. In order to maintain completeness, the algorithm should backtrack to all the possible inverted actions that could achieve the facts in the Pre-Processing Algorithm, even those with large application costs. However, due to the combinatorial explosion problem, this approach is not adopted by GRT.

A usual situation is the case where the sub-problems need further decomposition. This situation arises in two cases. The first is when two objects need each other to achieve their goals, as in the case of *grid* domain, with the keys and the robot, and the second case is when there is a sequential interaction between three or more objects. In these cases, the ordering graph of the initial problem encodes one aspect of the interaction, while the ordering graphs of the sub-problems encode other aspects. However, in order to avoid infinite decompositions, a cutoff level is defined.

## 6. The GRT Operation

GRT has been implemented in C++[5]. Its operation consists of several stages, which are shown in Figure 6a.

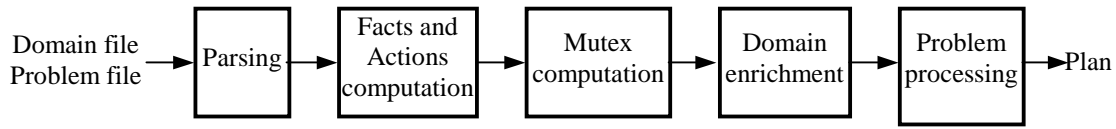

(a) The GRT operation stages

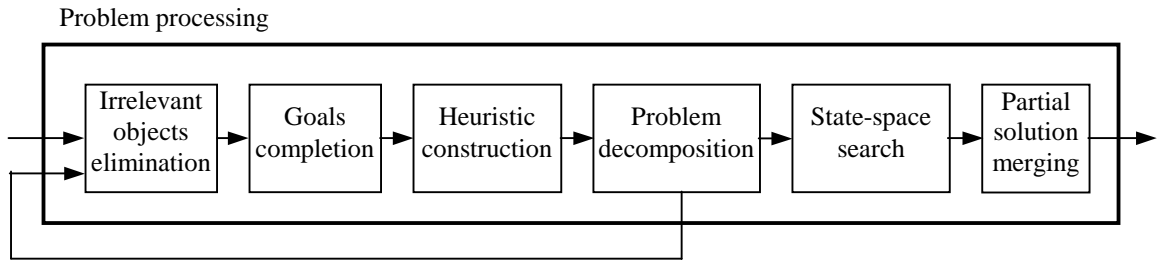

(b) The problem processing stage

Figure 6: The overall operation of the GRT planning system.

In the first stage the domain and problem files are parsed and the initial data structures are constructed. The second stage consists of computing the facts and the actions of the problem. The facts are stored in a tree structure, which is indexed by their predicates and their objects and allows for fast access, while the actions are stored in a linked list. Moreover, multiple pointers connect each fact with the actions, where the fact appears. The computation of the facts and actions is performed incrementally, by repeatedly applying the following steps:

- If a fact has been reached, create new actions that include this fact and others already reached, in their preconditions.
- If an action has been created, add its add effects in the tree structure.

The process starts with the initial state facts and continues until no more facts and actions can be reached. This approach is time efficient and succeeds in not generating many unreachable facts and actions. For example, in the *logistics* domain, the facts denoting that a truck is located in a city different than its initial location, and the corresponding actions, are not created. Note that in this stage, both the normal and the inverted actions are computed; the former are used in the mutex computation stage, while the latter are used in the heuristic construction stage. However, no pre-instantiated actions are used during the state-space search, where the applicable actions to each state are computed by progressively instantiating the action schemas, using constraint satisfaction techniques (forward checking and intelligent backtracking).

---

[5] GRT is available on-line at http://www.csd.auth.gr/~lpis/GRT/main.html.

The stages that follow are the computation of the mutual exclusion relations, the enrichment of the domain, and the problem processing. The latter stage consists of several sub-stages, as it is shown in Figure 6b, where the most important ones are the construction of the heuristic and the state-space search. Note that when we refer to the pre-processing phase of GRT, we mean all stages that precede the state-space search.

In the case where XOR-constraints are provided, GRT attempts to decompose the current problem into sub-problems. If this attempt is successful, the problem processing stage is executed recursively for each sub-problem, otherwise the current problem is solved. Finally, in the case of decompositions, the partial solutions are merged and the overall solution is returned.

## 7. Related Work

This section briefly presents other domain independent heuristic state-space planning systems, by emphasizing their similarities and differences to GRT, in terms of the way in which they construct their heuristic and the direction they traverse the state-space. We omit certain pieces of related work that concern specific pre-processing techniques implemented in GRT, as for example the elimination of irrelevant objects, since they have already been presented in previous sections.

The recent evolvement of the domain independent heuristic planning started with the work of Drew McDermott (1996, 1999) on UNPOP (UN-Partial Order Planner, UN- stands for non-). McDermott's planner is not restricted to pure STRIPS representations, supporting the more expressive language ADL (Pednault, 1989). The planner proceeds forward in the state-space. Distance estimates between states are based on the so-called regression graph, which is built from the goals using non fully-instantiated actions. UNPOP does not consider subgoals interactions and reconstructs the regression graph from scratch for each intermediate state. Although this planner is not competitive enough, compared to the subsequent heuristic planners, it was the faster one at the time of its appearance. However, we have to note that UNPOP has been developed in LISP, whereas the other heuristic planners are highly optimized C or C++ programs.

Although UNPOP was the first domain independent heuristic planner, the area has been pushed forward by the ASP (*Action Selection Planner*, Bonet, Loerings & Geffner, 1997) and HSP (*Heuristic Search Planner*, Bonet & Geffner, 1998) planners. The attractive feature of these planners is the simple way the heuristic is constructed, presented in Section 2.1. ASP used a best-first strategy with limited agenda, while HSP uses a hill-climbing one with limited plateau search and restarts (an in-depth presentation of the state-space search algorithms is given by Zhang, 1999).

Both ASP and HSP reconstruct their heuristic from scratch for each intermediate state. A variation, called HSPr (*r* stands for *regression*), constructs the heuristic only once (Bonet & Geffner, 1999). This approach resembles GRT, although HSPr constructs the heuristic forward and searches backwards. Both approaches have the problem of incomplete goal states, however it arises in different phases of the planning process. GRT faces this problem in the pre-processing phase, by enhancing the goals, as it has been described in Section 3. In HSPr, the problem arises in the search phase, in the form of invalid states in the regression state space. To cope with the problem, HSPr computes mutual exclusion relations and checks each state in the regression state space for any

REFANIDIS & VLAHAVAS

possible violation of these relations. The disadvantage of this approach is that it is considerably more time consuming than the GRT approach, since the HSPr has to check each visited state.

A variation of HSP, named HSP-2, changed the hill-climbing strategy to a best-first one, thus preserving completeness and producing better plans (Bonet & Geffner, 2001). Moreover, HSP-2 uses a weighted A* algorithm (WA*) (Pearl, 1983) of the form $f(S)=g(S)+W \cdot h(S)$, where $S$ is an intermediate state, $g(S)$ is the accumulated cost from the initial state, $h(S)$ is the estimated cost to reach the *Goals* and $W$ is a parameter. For $W=0$, the search algorithm behaves as a breadth-first one, for $W=1$ it behaves as the typical A* and for $W \rightarrow \infty$ it behaves as best-first. For the $h(S)$ function, HSP-2 supports several heuristic functions, apart from the one presented in Section 2.1.

Recently, two new planners, FF and ALTALT, appeared, which use a GRAPHPLAN-based approach to estimate distances between the intermediate states and the goals. ALTALT (*A Little of This*, *A Little of That*) is a regression planner based on HSPr, which faces the same problems with invalid states as HSPr (Nigenda, Nguyen & Kambhampati, 2000). ALTALT creates a planning graph in a pre-processing phase and uses several techniques to extract heuristic estimates of the distances between the intermediate states and the initial state. For example, one of them returns the level in the planning graph, where all the facts of the intermediate state appear, without any mutual exclusion relation between them.

FF (*Fast Forward*) is a forward heuristic planner (Hoffmann & Nebel, 2001). In order to estimate the distance between an intermediate state and the goals, FF creates a planning graph from the state to the goals, using relaxed actions. Since there are no delete effects, there are no mutual exclusion relations in the planning graph. From this graph, FF extracts a *relaxed plan*, the length of which is the distance estimate. Note that, since there are no mutual exclusion relations, no backtracking occurs during the extraction of the relaxed plan, thus the extraction is accomplished fast enough. The FF heuristic resembles the GRT one, in that both aim in obtaining under-estimates, but they adopt different approaches. The relaxations that FF performs are stronger, since it completely ignores the delete effects. So the FF estimates are usually smaller than the GRT's ones and most of the times are underestimates, whereas GRT not-rarely produces overestimates.

FF adopts a variation of the hill-climbing strategy, called *enforced hill climbing*, according to which, the planner always seeks to move to a state closer to the goals, according to its heuristic. FF achieves that by performing a bounded breadth-first search from the current state, with a maximum depth defined by the user; so the improving state does not have to be a direct successor of the current state. Once that an improving state is found, the new actions are added to the end of the current plan and the hill-climbing search continues from the new state. In the case where the bounded breadth-first search does not find an improving state, FF restarts the search from the initial state adopting a best-first search strategy.

FF exhibited distinguishable performance at the AIPS-00 planning competition. One of the features of FF resulting in its good performance is that it does not compute the applicable actions for each intermediate state. Actually, FF gives priority to the first level actions of the relaxed plan. Once that an action that produces a better state is found, it is applied and the next state is processed. Moreover, at most of the times, no new relaxed plan has to be constructed, since it suffices to remove the lastly applied action from the beginning of the previous relaxed plan. So, FF succeeds in reducing drastically the cost of processing each intermediate state, paying however the cost of loosing completeness.

The bottleneck that occurs while determining the applicable actions for each intermediate state has also been identified by Vrakas et al. (1999, 2000). In this work, the process of finding and applying the applicable actions has been parallelized, resulting in almost linear speedup. Parallelizing the process of finding the applicable actions, instead of ignoring most of them, as FF does, presents the advantage of preserving completeness; however, the cost is that a parallel machine is required.

We close the reference to other heuristic state-space planners with the STAN planning system (Fox & Long, 1998; Long & Fox, 1999). STAN is not a heuristic state-space planner, at least in its basic architecture, but a graph-based planner, which uses several pre-processing techniques for extracting useful domain information that is exploited for more efficient graph construction and solution extraction. However, in the AIPS-00 competition a hybrid architecture was used (Long & Fox, 2000; Fox & Long, 2001), where a heuristic state-space planning module was employed to solve specific identified sub-problems. Thus STAN succeeded in improving its performance, especially in cases of transportation domains.

Concerning problem decomposition, work has been done on goal ordering (Cheng & Irani, 1989; Drummond & Currie, 1989). Recently a similar approach has been proposed by Koehler (1998) and has been extended by Koehler and Hoffmann (2000). This approach automatically derives an ordering relation between the goal facts, which can be used by any planner to search for increasing sets of subgoals. The advantage of this approach is that no extra information is needed, except for the usual domain definition, while the disadvantage, with respect to the XOR-constraints approach, is that only the goal facts are taken into account in the intermediate states that are constructed. This approach has been adopted by the FF planning system.

# 8. Performance Results

In this section, we present performance results from several domains, taken from the literature and from the two planning competitions. First, we investigate how the several techniques of GRT contribute to its overall performance and then we compare GRT to other planners.

The measurements that follow were taken on a SUN Enterprise 3000 machine running at 167MHz, with 256 MB main memory and operating system Solaris 2.5.1. In the experiments we set a 5 minutes time limit for all experiments and planners[6].

## 8.1 Measuring the Effectiveness of the Related Facts

In order to measure the contribution of the related facts to the overall performance of GRT, we tested the planner, with and without related facts, on problems from various domains. The results (solution length and time) are presented in Figure 7 (a-f).

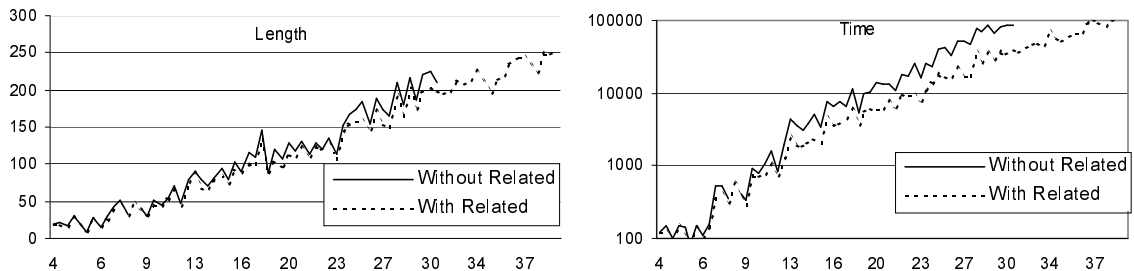

(a) *Logistics* problems (the goals have been enhanced with the *most promising* facts selection method)

---

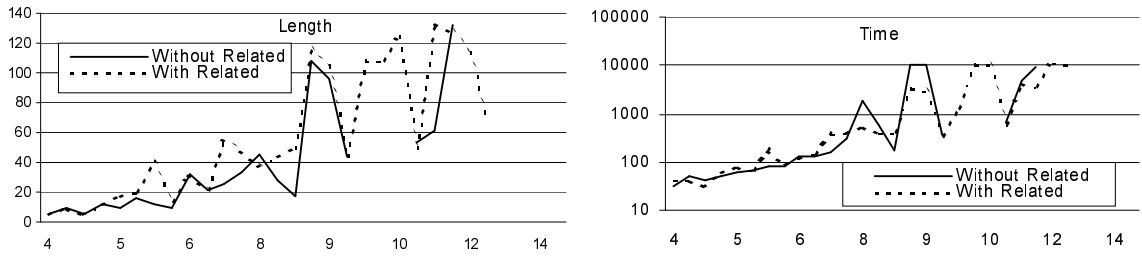

(b) *Blocks-world* problems with 4 action schemas (*push, pop, pick-up, put-down*)

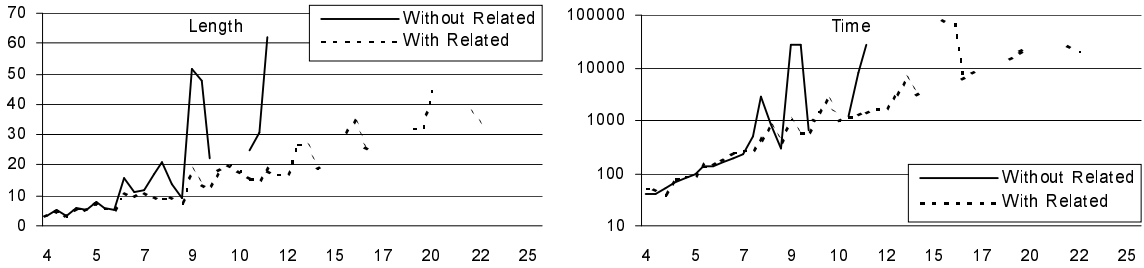

(c) *Blocks-world* problems with 3 action schemas (several cases of *move*)

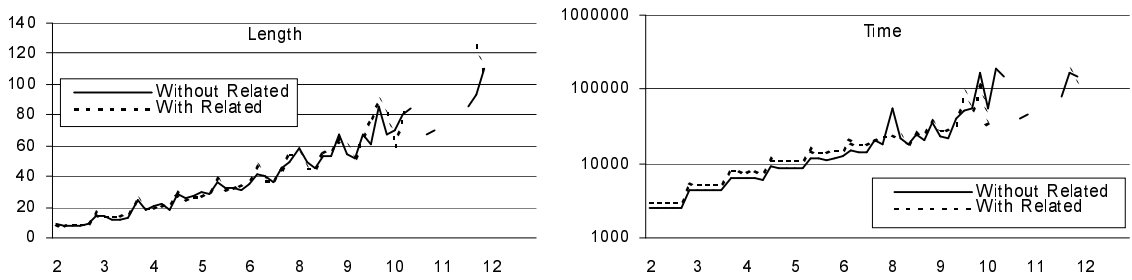

(d) *FreeCell* Problems

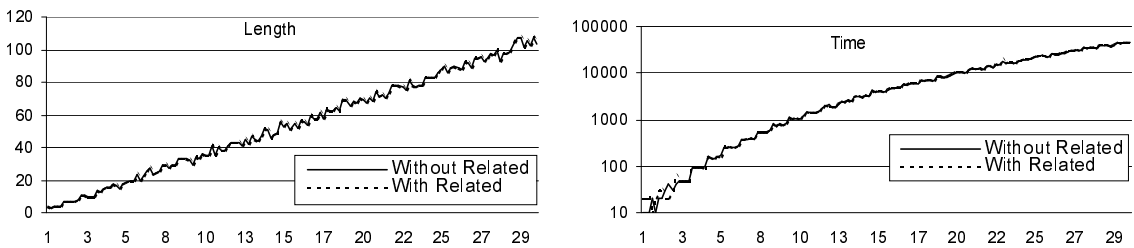

(e) *Elevator* problems

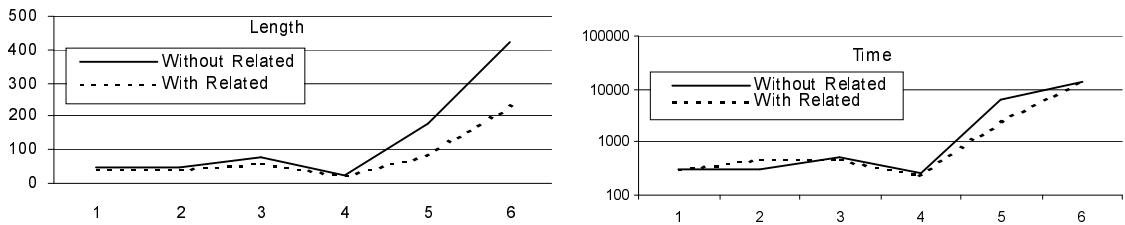

(f) *Puzzle* problems

Figure 7: Solution length and time (in msecs) with and without the use of related facts for problems from several domains.

We can classify the above domains in three groups. The first group includes the domains where the use of related facts clearly improves both the solution length and time. This group comprises the *logistics* domain (6a), the *blocks-world*, when a 3-action schemas representation (*move* actions) is used (6c), and the *puzzle* domain (6f). In these domains, there were many cases where GRT without related facts did not solve the problems, while with the related facts it did. Moreover, in most cases when both versions solved a problem, the version with the related facts was faster and came up with a shorter plan.

The second group includes domains where the use of related facts does not affect the effectiveness of the planning process. This group comprises the *elevator* domain, along with the *gripper*, the *movie* and the *mystery* ones. In these domains, there is usually a single way to achieve the goals, so both versions produce identical plans. However, due to the processing overhead, imposed by the computation of the related facts, the version with the related facts is slightly slower than the version without them.

Finally, the third group includes the domains where there is no apparent predominance between the two versions. The *freecell* domain and the *blocks-world* domain, when a 4-action schemas representation is used (*push*, *pop*, *pick-up*, *put-down*), fall into this class. In these domains the two versions do not have equal performance, but there are problems where one version surpasses the other and vice-versa.

The conclusion drawn from the above measurements is that the effectiveness of the related facts depends on the domain. They are more suitable in domains where there are several ways to achieve the goals, as *logistics* or *blocks-world*.

Additionally, their efficiency depends on the way the domain is codified. A typical example is the *blocks-world* domain and the 4- and 3-action schemas representations. The problem with the 4-action schemas representation is that *pushing* and *stacking* a block anywhere has always the same fact as precondition, i.e. that the block is held by the arm. The consequence is that neither the related facts, nor the distances are computed correctly. However this is not a problem of the related facts, it is a common problem in domain independent heuristic planning, as it results from the last planning competition. On the other hand, if a 3-action schemas representation is used, then the paths to achieve the facts of the domain are better tracked, so larger problems can be solved and the contribution of the related facts is significant. We believe, finally, that also in the *freecell* domain there is a representation inefficiency, however we have not yet tried to construct an alternative one.

## 8.2 Using Several Methods to Enhance the Goals

In order to measure the effectiveness of the three proposed methods to enhance the goals, we ran GRT using them in the *logistics* problems of the AIPS-00 competition. We selected this domain, since in the other domains of the competition the goal state is either complete, or near complete, so there is no difference among the three methods. Figure 8 shows the solution length and time for the easiest of the *logistics* problems.

With regard to solution length, the first method, which considers all the candidate facts as goal facts, always came up with better plans. As we mentioned in Section 3.2, this method produces small differences among the estimated distances, so the search process tends to be breadth-first. However, in most of the cases, the third method found plans of equal quality. With regard to the solution time, the last two methods work faster, since they produce greater differences between the distances.

In Section 3.3 we also presented a method of enriching the domain representation. As already mentioned, we were motivated by the need to treat domains like the *movie* or the *elevator*. We do not present comparative performance results between the domain enrichment method and the pure GRT planner for these domains, since without this technique it is impossible for GRT to solve the problems. However, it would be interesting to test the efficiency of this method to other heuristic state space planners.

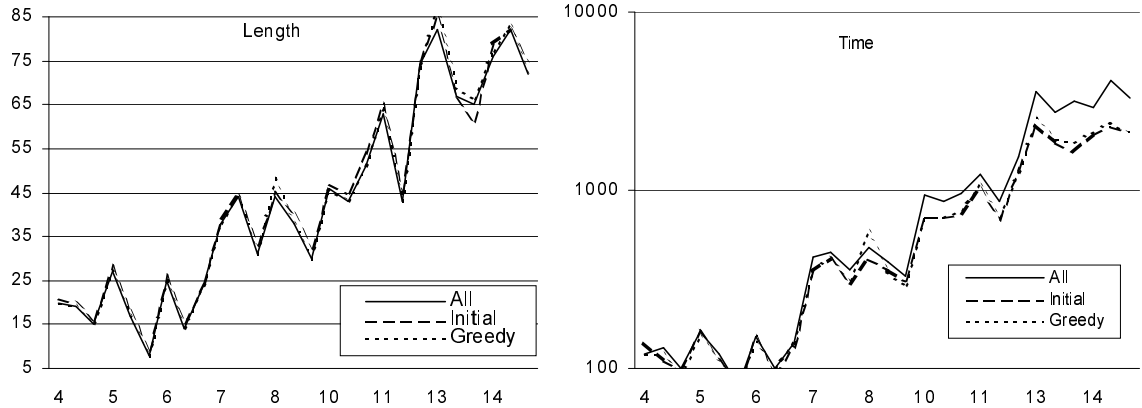

Figure 8: Results for *logistics* problems using different methods to complete the goals.
All = Consider all the candidate facts as goal facts.
Initial = Select the initial state facts.
Greedy = Favor the most promising facts.

## 8.3    Reducing the Size of the Problem

The work of detecting and eliminating irrelevant objects has been motivated by the need to simplify the sub-problems resulting after the decomposition of a problem, when using XOR-constraints. Performance results for this case are presented in Section 8.4. This section presents indicative results concerning the effectiveness of the technique in the *colored logistics* domain that has been mentioned in Section 4.1. For this purpose we enhanced the first group of *logistics* problems of the AIPS-00 competition with the required predicates and actions and we added propositions defining the original color of each package to the initial states. Figure 9 presents the time needed to solve the problems, with and without the irrelevant objects elimination technique. As it results from the experimental data, there is an improvement in the solution time of about 20%. Note that in both cases the same plans have been found; however, this would probably not be the case in other domains.

In order to measure the efficiency of the numerical representation of resources, we ran GRT both in the original *mystery* domain and in a modified domain, where resources have been represented by numbers. Figure 10 presents the time needed to solve the problems with both cases of GRT. Note that in these experiments only the solvable *mystery* problems have been taken into account. As it results from Figure 10, GRT was significantly faster, when a numerical representation is used. The improvement is 65% on average. As for the solution length, in both cases the same have been found again.

Both techniques evaluated in this section gain their speedup mainly from the pre-processing phase, since distances for a significantly smaller number of facts have to be estimated. As for the search phase, there is also a speedup, but is less important. Actually, the number of applicable actions to each state is the same with the two alternative representations of resources, since these are equivalent. Moreover, the detection of the applicable actions in the atom-based representation takes about the same time, due to the effective constraint-satisfaction techniques that GRT uses when instantiating the action schemata. Concerning the elimination of irrelevant objects, without this technique, there are more applicable actions to a state, which however are usually not selected, since they do not lead to an improving state. However, the time spent in the detection of these actions may be not negligible.

The significance of the two techniques lies in that the overall time needed to solve the problems remains about the same, in the case where more irrelevant objects are used, and exactly the same, in the case where more resource levels are used. In the case of more irrelevant objects, these are detected (in negligible cost) and eliminated from the subsequent stages (Figure 6). However, there is some overhead imposed by the stages that precede the irrelevant objects elimination stage, from where these objects have not been eliminated.

In the case of more resource levels, these do not lead to the generation of new ground facts and actions, so all the pre-processing stages consume exactly the same time. As for the state-space search, this is also executed in the same time, but only in the case where neither the initial availability of resources, nor their consumption by the actions, nor finally the constraints over them have been changed. If this is not the case, then we are dealing with a different planning problem, which may be harder to solve.

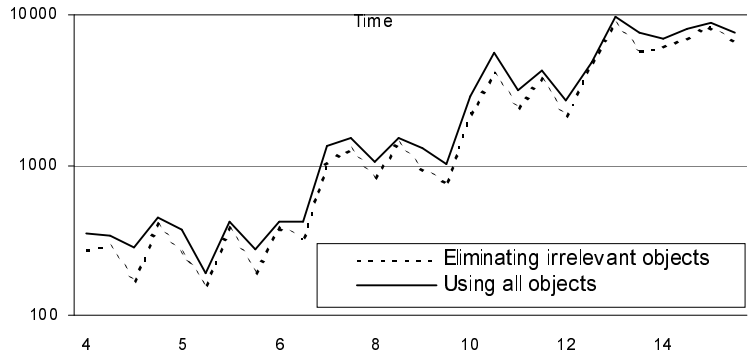

Figure 9: Time (in msecs) needed to solve the *colored logistics* problems, with and without the irrelevant object elimination technique.

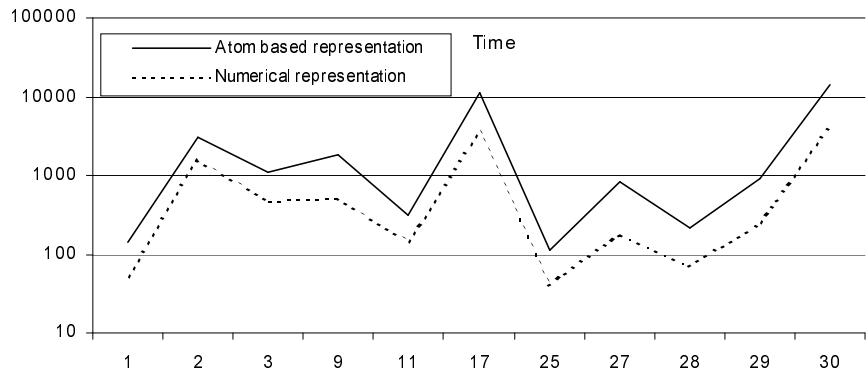

Figure 10: Time (in msecs) needed to solve the solvable *mystery* problems, when the original atom-based or a number-based representation for resources is used.

## 8.4 XOR Constraints

We tested the efficiency of the XOR-constraints based decomposition in two domains: A *simplified mystery* domain, where resources have been removed, and the *grid* domain of the AIPS-98 competition. We did not use the *logistics* domain for these experiments, since *logistics* problems are not difficult for the original GRT and the small profit from solving the easier sub-problems is compensated by the extra pre-processing cost of each sub-problem.

We removed resources from the original *mystery* domain because otherwise it would be probable to obtain unsolvable subproblems. As it has been noted in Section 5, decomposing a problem may lead to loss of completeness, thus the technique is unsuitable for domains where deadlocks may arise, as the original *mystery* one. Note that by removing resources, all *mystery* problems become solvable.

The XOR-constraints that have been defined for the *simplified mystery* domain were the following:

> ( ( *xor* ( *at ?Truck* * ) ) (*truck ?Truck* ))
>
> ( ( *xor* ( *at ?Package* * ) (*in ?Package* * ) ) ( *package ?Package* ) )

while for the *grid* domain were the following ones:

> ( ( *xor* ( *at-robot* * ) ) )
>
> ( ( *xor* ( *locked ?Place* ) ( *open ?Place* ) ) ( *place ?Place* ) )
>
> ( ( *xor* ( *at ?Key* * ) ( *holding ?Key* ) ) ( *key ?Key* ) )

Note that in the grid domain an XOR-constraint denoting that the arm is either empty, or the robot holds a key has not been defined, since this would lead to pointless decompositions.

In both domains, we ran GRT with and without the problem decomposition technique. Additionally, in order to demonstrate the contribution of the irrelevant objects elimination technique when solving the sub-problems, we conducted experiments for this case in the *simplified mystery* domain. We did not consider this case in the *grid* domain, because no irrelevant objects can be detected there. Figure 11 presents the results.

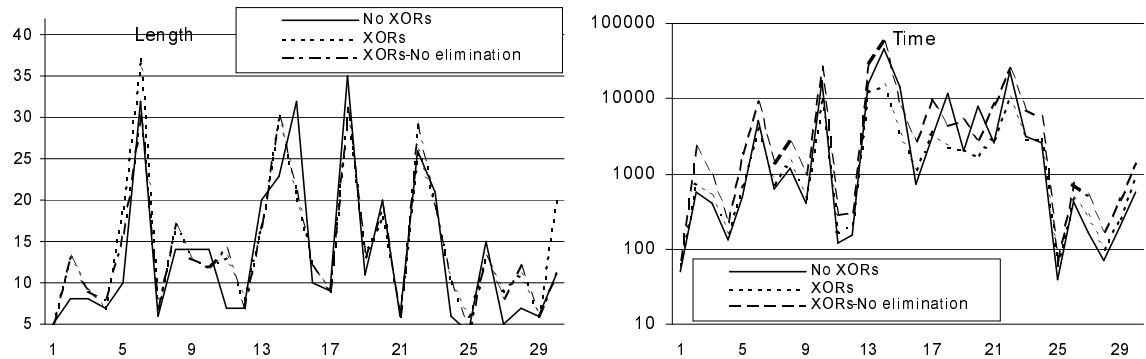

(a) *Simplified Mystery*

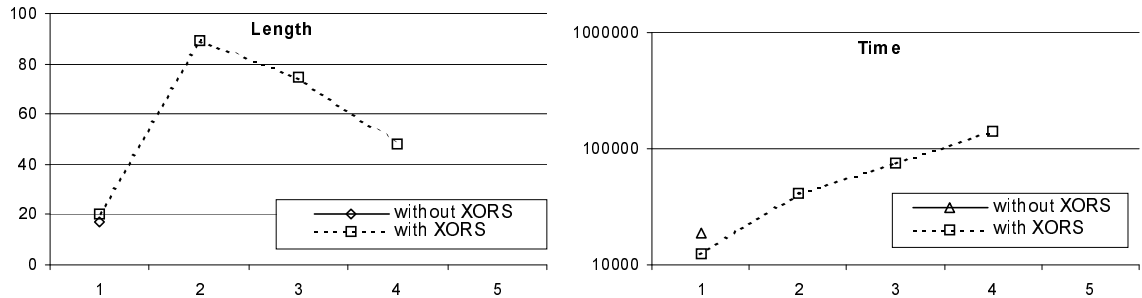

(b) *Grid* Domain

Figure 11: Solution time (in msecs) and length with and without the XOR-constraints based problem decomposition technique.

As for the simplified mystery domain, GRT without the problem decomposition technique generally produced shorter plans, as expected. On the other hand, the use of the XOR-constraints accelerated the problem decomposition process, especially in case of difficult problems. Actually, if we only consider the seven most difficult problems, the improvement achieved by the decomposition is 60% on average. Note however that, when the irrelevant objects elimination technique was not used, there was no improvement. In not difficult problems there is no acceleration, since, as in the case of the *logistics* problems, the small profit from the faster solution of the easier sub-problems is compensated by the cost of repeating the pre-processing phase for each one of them.

The *grid* domain was the most difficult one of the AIPS-98 competition. The contestants managed to solve only the first problem. GRT without XOR-constraints could only solve the first problem, too. On the other hand, with the XOR-constraints based decomposition, GRT was able to solve the first four problems in the time limit of 5 minutes, while in the fifth problem it ran out of memory. It is worth noting that this domain produces multiple levels of decompositions. Figure 12 presents these levels for the strips-grid-y-2 problem.

As far as we know, the only planner that can cope with the *grid* problems effectively is FF. We ran FF in the five *grid* problems and it solved the first four, within the time limit of 5 minutes, with the following results (length/time): 14/230, 39/840, 40/7810 and 45/3280, which are considerably better compared to the performance of GRT.

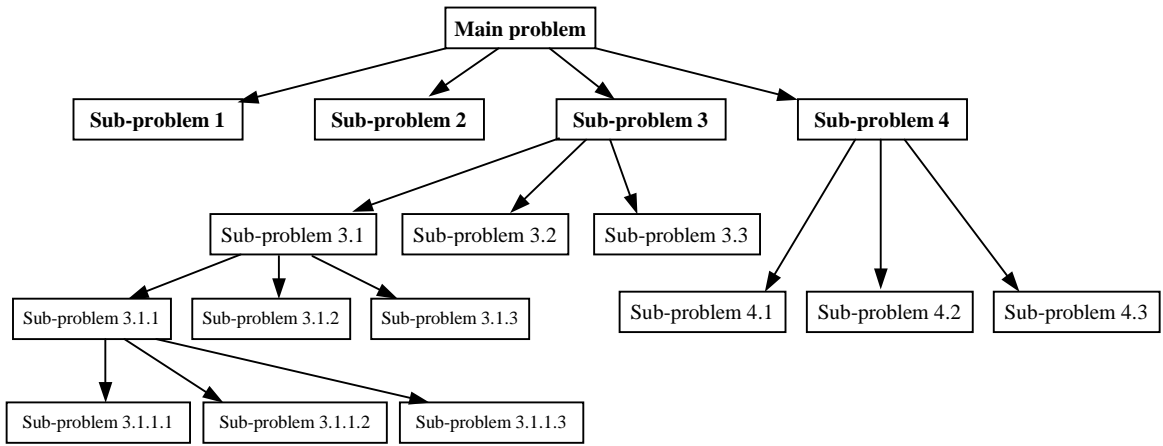

Figure 12: Decomposition for the strips-grid-y-2 problem using XOR-constraints.

## 8.5 Best-First and Hill-Climbing Strategies

Recently we equipped GRT planner with two new features: a second optional search strategy, the well known hill-climbing, and a closed-list of visited states, in order to avoid revisiting them.

GRT adopts the *enforced hill-climbing* strategy, originally presented in Hoffmann & Nebel (2001), according to which, from each intermediate state a limited breadth first search is performed, until an improving state is reached. When an improving state cannot be found, GRT restarts the search from the initial state with the typical best-first strategy.

Moreover, the hill-climbing strategy has been enhanced with a fast action selection mechanism. As it has been presented in Section 5.3, when GRT estimates the distances between the problem's facts and the goals in the pre-processing phase, it stores in the GRG structure the action that achieved each fact. So, in order to find an improving successor state quickly, the hill-climbing search strategy first attempts to apply the actions that achieved the current state's facts. Once that an improving successor state is found, the remaining of the actions are not processed, thus avoiding to compute all the applicable to the current state actions. Note however that it is not guaranteed that these actions can always be applied to the current state. In case where no improving state can be found, the remaining of the applicable to the current state actions are taken into account.

Figure 13 presents comparative performance results in *logistics* and *elevator* problems, using both search strategies. In the *logistics* problems, the *most promising facts* selection method of enhancing the goals has been used. As it results from the experimental data, in the *logistics* problems and with the use of the hill-climbing strategy, there is a significant reduction in the solution time of about 52%. The cost is an increment of about 3% in the length of the plans. In the *elevator* problems, there is also a reduction in the solution time of about 29%, whereas the produced plans are identical.

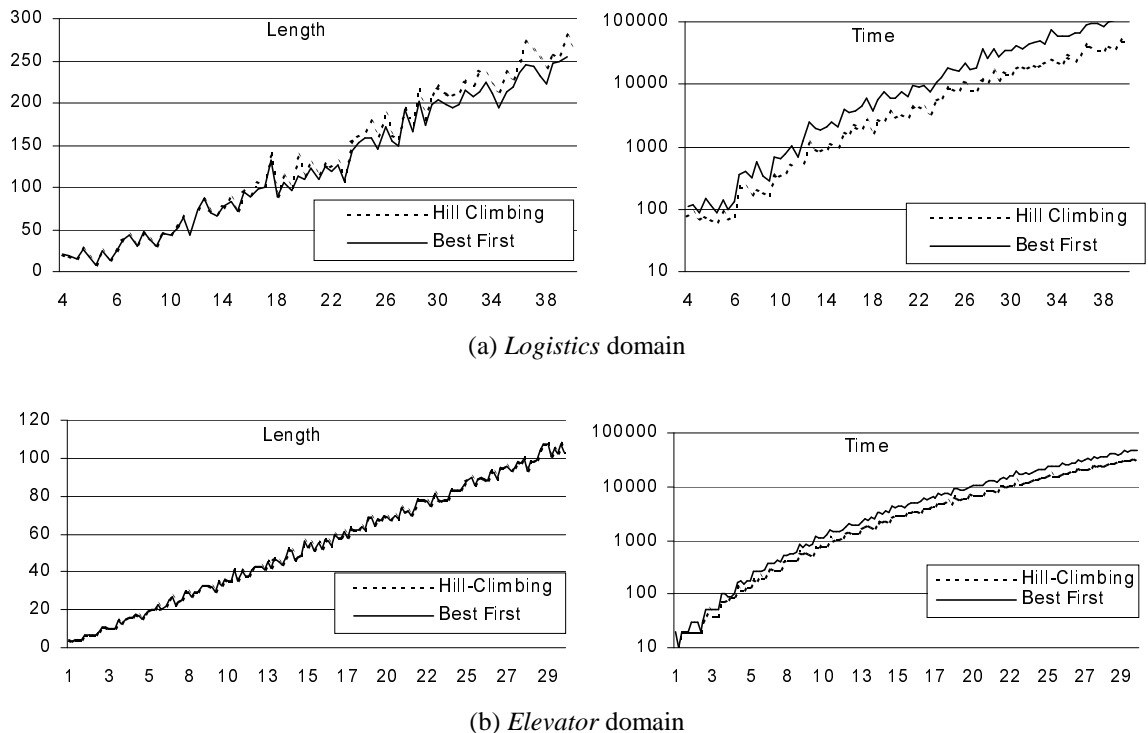

(a) *Logistics* domain

(b) *Elevator* domain

Figure 13: Comparative results (solution length and time) between the hill-climbing and the best-first strategies.

We tested the efficiency of the fast action selection mechanism, by also running GRT with the hill-climbing strategy but without this mechanism in the same *logistics* and *elevator* problems. Concerning the *logistics* problems, the speedup was about 47%, while the increment in the solution length was 3% on average again. Concerning the elevator problems, the speedup was 28%, whereas the produced plans were again identical. The conclusion from these additional measurements is that the speedup is primarily due to the hill-climbing strategy and secondly due to the fast action selection mechanism. The contribution of this mechanism depends on the domain and it is more important in the *logistics* and less in the *elevator*. Its inefficiency in the *elevator* domain means that the actions that are selected by this mechanism do not usually lead to an improving state or they are not applicable, so all the applicable actions have to be computed.

Results for other domains, like *blocks-world* and *freecell*, are not presented, since in these domains hill-climbing usually fails to find a plan and GRT restarts on a best-first basis. However, in these domains the closed-list of states has been proved invaluable, improving drastically the performance of GRT. For example, in the *freecell* domain and without the closed list of visited states, the GRT planner in the AIPS-00 planning competition succeeded in solving problems with up to 6 cards per suit, while with this data structure it can solve some of the more difficult ones (13 cards per suit). Note that for an efficient implementation of the closed-list of visited states a hash-table data structure has been adopted.

## 8.6 Comparison to other Planners

In this section, we present comparative results between the GRT planner and other planners. We decided to use HSP-2 (Bonet & Geffner, 2001), FF (Hoffman & Nebel, 2001), STAN (Long & Fox, 2000; Fox & Long, 2000, 2001) and ALTALT (Nigenda, Nguyen & Kambhampati, 2000)[7]. All these planners took part in the domain independent track of the AIPS-00 planning competition. We selected these planners because HSP-2 and STAN are state-of-the-art planning systems, FF has been awarded for its outstanding performance in the last competition and ALTALT is a new but very promising domain-independent state-space heuristic planner.

The aim of our experiments is to have an overall view of the performance of the evaluated systems. Performing pair wise comparisons between specific optimization techniques is not possible, since these techniques are implemented on top of different systems. Moreover, this kind of comparisons is out of the scope of this paper, which focuses in the use of specific directions for constructing the heuristic and traversing the space of the states, in the area of domain-independent heuristic state-space planning, and not in the evaluation of the numerous pre-processing optimization techniques. However, in the cases where we identify the contribution of a specific feature in the performance of a planner, we comment on this.

In order to have fair comparisons, we used exactly the same problem and domain description files for all planners. So, GRT ran without XOR-constraints or numerical representation of resources. Moreover, although the irrelevant object elimination technique is an integral feature of GRT, it had no contribution in these domains, since there were not irrelevant objects. We believe that the absence of irrelevant objects in these domains does not mean that this technique has limited applicability, but it is an indication that more real domains for testing purposes have to be used in the future, since the planning tasks in our real-life are full of irrelevant objects. Finally, the domain enrichment technique proved valuable for the *elevator* domain only. However, this technique, as well as the goal enhancement one, has not to be seen as an optimization technique, but as a way to overcome the problems that arise from the backward direction of the heuristic construction.

We tested the planners in several domains taken from the planning competitions and from the literature, in the same workstation and within the 5 minutes time limit. The results are presented in the following.

### 8.6.1 LOGISTICS

For the *logistics* domain we used the test suite of the AIPS-00 competition. The results are shown in Figure 14. In this domain GRT, as well as FF and STAN, performed well, solving all the problems. HSP and ALTALT failed to solve the large problems within the time-limit. In general, best plans are found by STAN, which uses special domain-dependent heuristics for problems identified as

---

transportation problems. Best solution times are achieved by FF and STAN in the small problems and by GRT in the large ones.

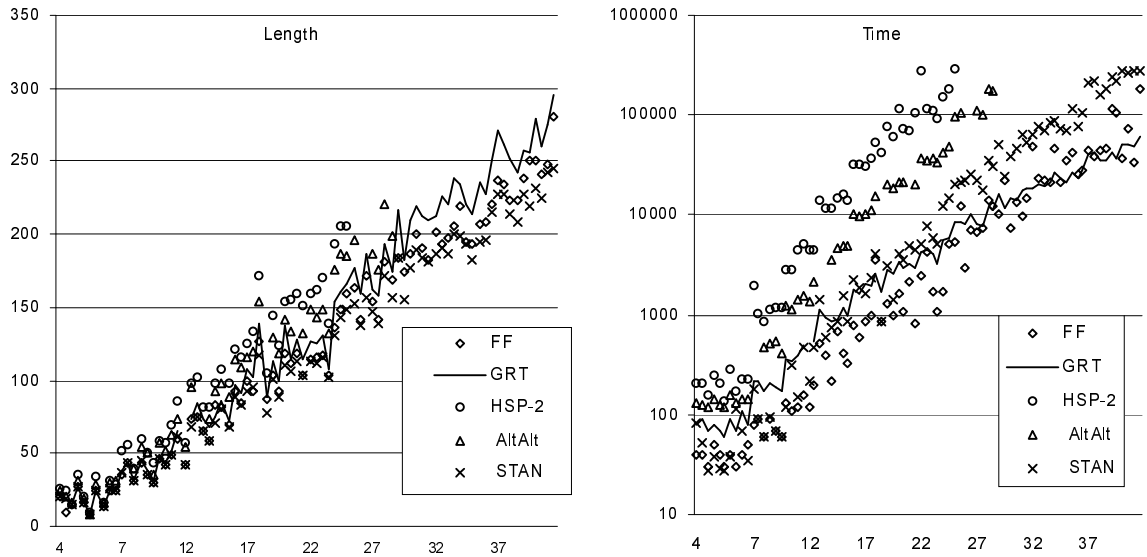

Figure 14: Solution length and time (msecs) for the *logistics* problems of the AIPS-00 competition.

The *logistics* problems in Figure 14 have incomplete goal states. GRT ran with the *most promising facts* goals-completion method and with the hill-climbing strategy. However, the incompleteness of the goal state is an advantage for the planners that construct the heuristic in a forward direction. Motivated by this remark, we forced all the planners to solve *logistics* problems with complete goal states, requiring all the trucks and planes to return to their initial location. The results are shown in Figure 15.

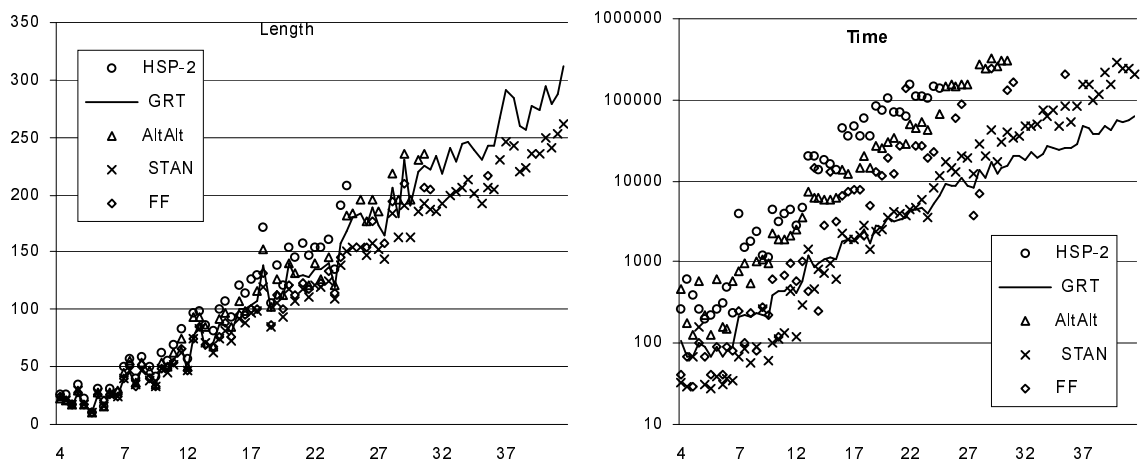

Figure 15: Solution length and time (msecs) for *logistics* problems with complete goal states.

In the new *logistics* problems, GRT, STAN and HSP-2 exhibited stable performance, solving the problems in about the same time. For GRT, this means that the goal completion mechanism behaves well, at least in this domain. FF failed to solve the large problems. Finally, ALTALT solved some more problems and this is because the regression mechanism did not encounter invalid states. Note that, although the goal state was complete in this case, GRT treated these problems as usual, attempting to enhance the goals.

### 8.6.2 BLOCKS-WORLD

For *blocks-world* problems in the AIPS-00 competition a four-actions representation was used, i.e. actions *push*, *pop*, *stack* and *unstack*. This representation is unsuitable for GRT, as it has been explained in Section 7.1. So, GRT did not solve most of the *blocks-world* problems. Figure 16 presents the results of all planners in all *blocks-world* problems.

As shown in Figure 16, FF exhibits the best performance, solving the majority of the problems and producing better plans than the other planners. The superiority of FF in this domain is due to a technique called *Added Goal Deletion*, according to which the goal facts are ordered and achieved in a progressive manner (Hoffmann & Nebel, 2001; Koehler and Hoffmann, 2000). This technique is especially suited for the *blocks-world* domain and the 4-action schemas representation. However, this technique does not always succeeds to produce good orderings and this is the reason why FF fails to solve some of the easiest problems, which have been solved by the other planners.

As for the remaining planners, HSP-2 succeeded in solving all problems with up to 18 blocks and one problem with 24 blocks, GRT and ALTALT solved problems up to 14 blocks and STAN up to 12 blocks. Moreover, GRT produced plans of low quality.

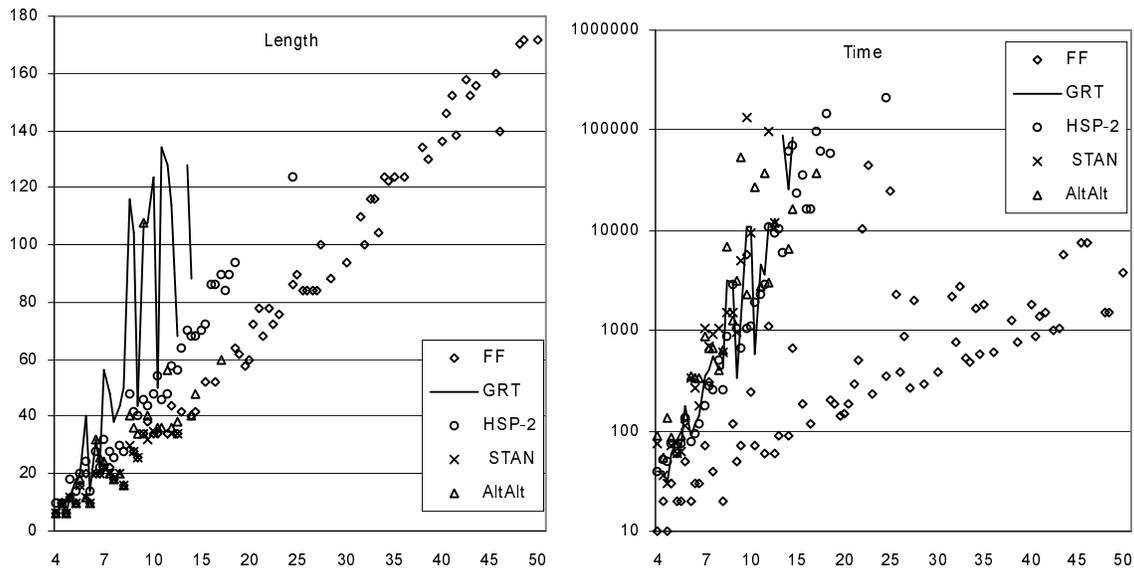

Figure 16: Solution length and time (msecs) for the *blocks-world* problems
using the 4-action schemas domain representation.

In order to demonstrate the influence of the domain representation in the efficiency of GRT, we ran all the planners in the same problems using the alternative 3-action schemas domain representation. The results are shown in Figure 17.

The performance of GRT is significantly improved, solving problems with up to 33 blocks and producing better plans than the other planners. Moreover, with the exception of the smallest problems, GRT is faster than the other planners, but FF. The latter solved less large problems, but solved all the smallest ones. HSP-2 solved all the problems with up to 19 blocks, while ALTALT and STAN stopped at 14 blocks.

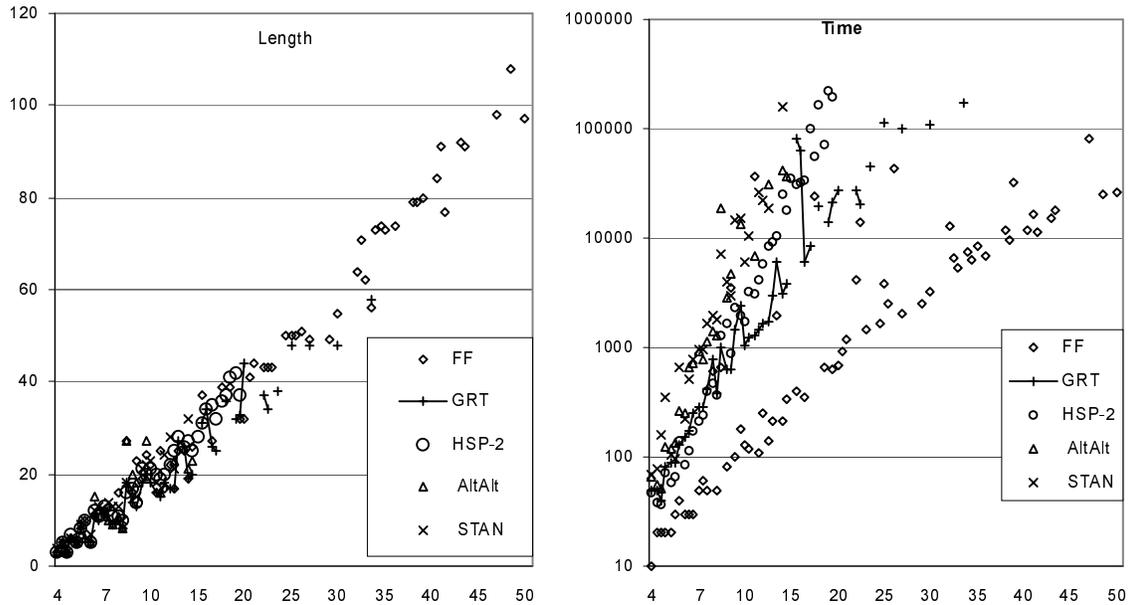

Figure 17: Solution length and time (msecs) for the *blocks-world* problems using the 3-action schemas domain representation.

### 8.6.3 FREECELL

*Freecell* is the famous card game taken from the MS-Windows 98 distribution. The domain was initially introduced in the AIPS-00 competition and proved one of the most difficult domains. Figure 18 presents the performance results in this domain. Note that ALTALT could not solve these problems and this was also the case in the competition.

In the *freecell* domain, the only planners that succeeded to solve some of the difficult problems were GRT and FF. Actually, these planners solved some problems with 12 and 13 cards per suit. HSP-2 solved problems with up to 6 cards per suit and STAN up to 3 cards per suit. Regarding the solution quality, GRT produced better plans than FF. Regarding the solution time, FF was faster in the small problems, whereas in the big ones the two planners had equal performance.

### 8.6.4 ELEVATOR

The *elevator* (or *miconic-10*) domain has been presented in Section 3.3. At least in its pure STRIPS version, it is a relatively easy domain. So, all planners found plans of equal quality (with the exception of HSP-2, which produced slightly more lengthy plans). However, the planners have different performance in terms of solution time.

Specifically, FF was the fastest, followed by STAN, then GRT, then HSP-2 and finally ALTALT. This domain favors FF, because the relaxed plan produced by its heuristic mechanism for the initial state is actually the solution, since the original actions of the domain do not contain any delete lists. STAN identifies this domain as a transportation domain and uses suitable techniques to solve the problem. Finally, GRT is faster than HSP-2 and ALTALT, since GRT constructs its heuristic faster than HSP-2. The results are presented in Figure 19.

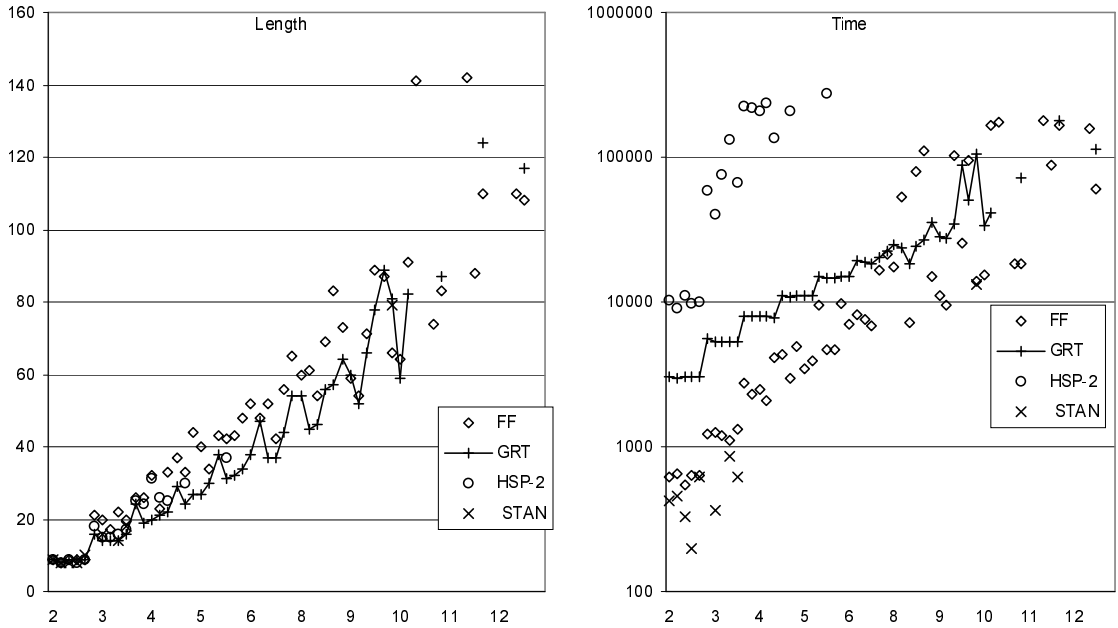

Figure 18: Solution length and time (msecs) in the *freecell* domain.

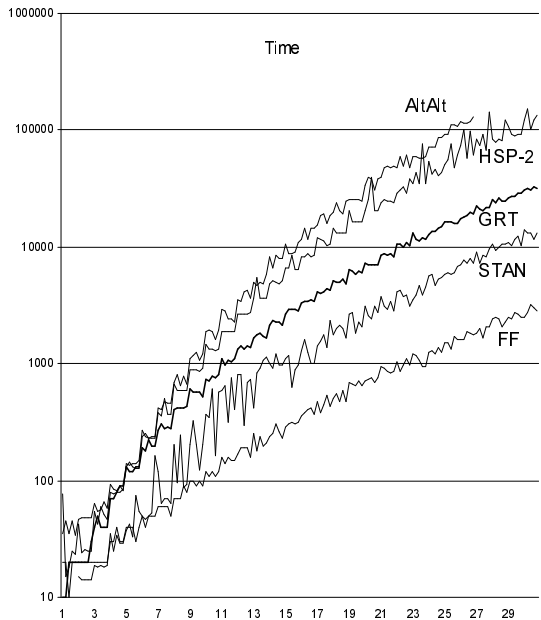

Figure 19: Solution time (in msecs) in the *elevator* domain.

### 8.6.5 Gripper

The *gripper* domain was introduced in the Aips-98 planning competition. The domain concerns a robot with two grippers that must transport a set of balls from one room to another. In the Aips-98 competition, only Hsp managed to solve the 20 problems. Figure 20 presents the results in this domain.

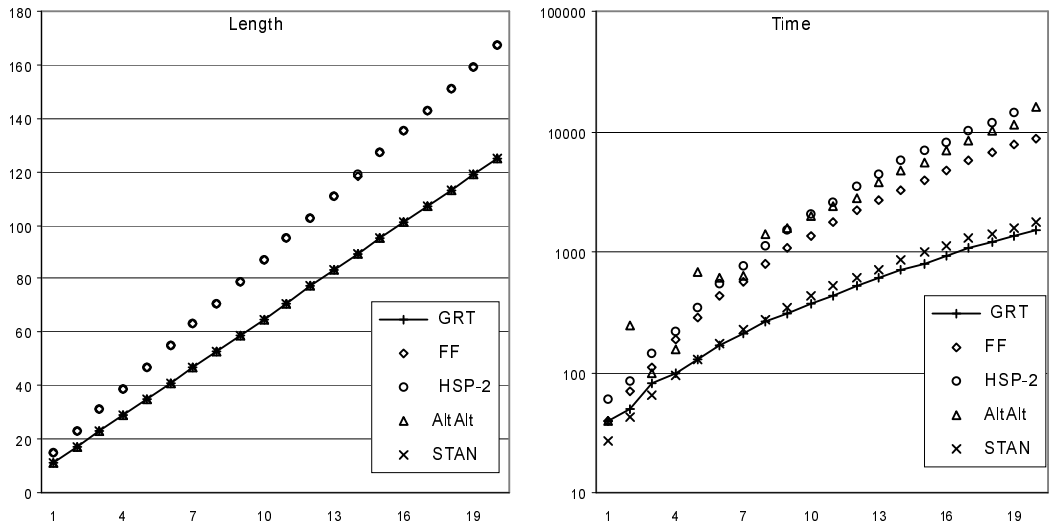

Figure 20: Solution length and time (msecs) in the *gripper* domain.

Regarding the solution length, the five planners have been divided into two groups: GRT, ALTALT and STAN produced identical plans of higher quality, while FF and HSP-2 produced identical plans of lower quality. Regarding solution time, GRT is the fastest planner in all problems apart from some of the easiest, followed closely by STAN, next comes FF, next ALTALT and last HSP-2. Note that in this domain STAN takes advantage of its symmetry analysis, which identifies the set of the balls and the two grippers as symmetric objects (Fox and Long, 1999).

### 8.6.6 HANOI

We ran the planners in 6 *hanoi* problems, taken by Bonet and Geffner (2001). The six problems have three to eight disks respectively. Figure 21 presents the results.

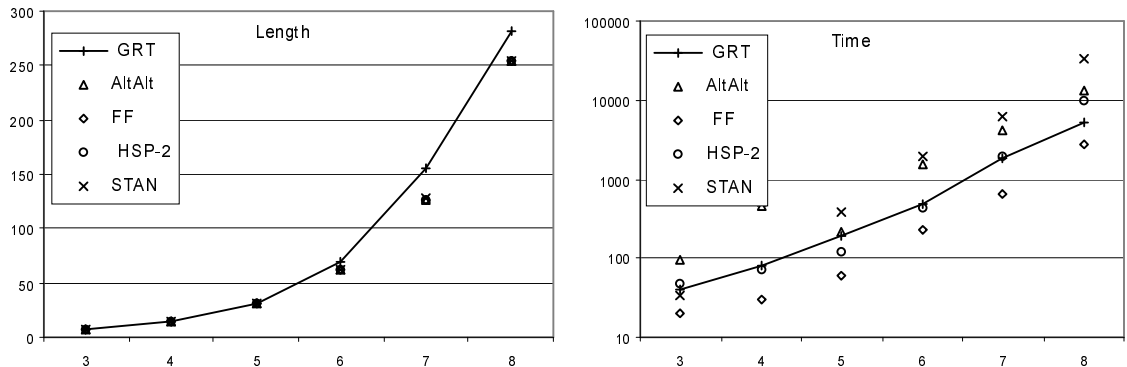

Figure 21: Solution length and time (msecs) in the *hanoi* domain.

Regarding the solution length, all the planners found identical plans, with the exception of the last two problems, where GRT found worse plans. Regarding the solution time, FF was the faster, then came GRT and HSP-2, then ALTALT and last came STAN.

8.6.7    PUZZLE

We ran the planners in four 8-puzzle problems and in two 15-puzzle ones, taken by Bonet and Geffner (2001). Two of the four 8-puzzle are hard and their optimal solution involves 31 actions, the maximum plan length in this domain. The 15-puzzle problems are of medium difficulty. Figure 22 presents the results.

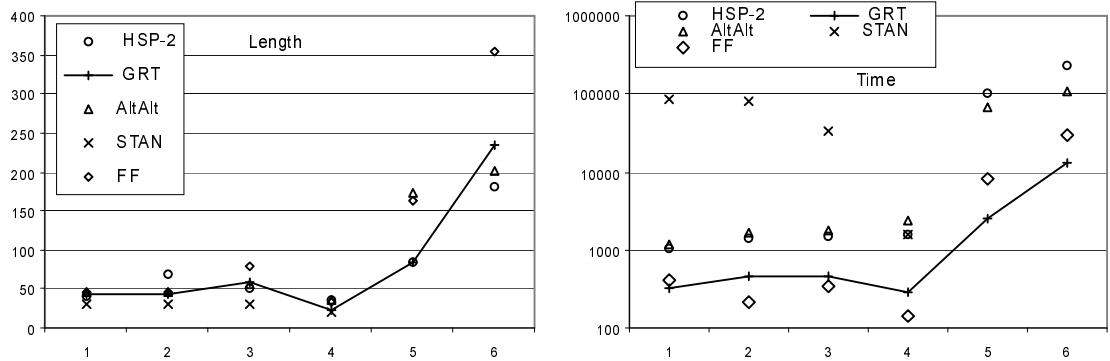

Figure 22: Solution length and time (msecs) in the *puzzle* domain.

STAN solved only the 8-puzzle instances, but it produced the best plans. The other planners solved all the problems, but they presented variations in the quality of their plans, with the FF planning system producing worst plans in most of the cases. Regarding solution time, FF was the fastest in the easier problems and GRT in the more difficult ones, followed by HSP-2 and ALTALT. STAN was the slowest planner in this domain.

## 9. Conclusion and Future Work

In this paper we presented the GRT planning system, a heuristic state-space planner, which constructs its heuristic in a domain-independent way. The fundamental difference between GRT and other heuristic state-space planners is that GRT constructs its heuristic once, in a pre-processing phase and in a backward direction, using regression from the goals. GRT attempts to track the positive and negative interactions that occur between the problem facts when trying to achieve them, in order to produce better estimates.

GRT employs several new techniques that improve its efficiency. These are the automated identification of incomplete goal states, the identification and enrichment of inadequate domain representations, the elimination of irrelevant objects and the adoption of a numerical representation of resources. Finally, a knowledge-based method that uses domain axioms in the form of XOR-constraints, in order to decompose difficult problems into easier sub-problems that have to be solved sequentially, has adopted.

The paper presented extensive comparative results in a large number of domains. In the comparisons, besides GRT, four of the most powerful domain independent planners took part. The results showed that no planner clearly outperforms all the others.

Concerning solution time, in most of the domains GRT and FF were the fastest planners. The explanation behind this observation lies in that these planners construct their heuristic either once (in the case of GRT), or a few times only (in the case of FF). For example, in the *elevator* domain, where delete effects do not exist and FF constructs a relaxed planning graph only once, it is extremely fast. On the contrary, in the *gripper* and the *puzzle* domains, where FF needs to reconstruct the relaxed planning graphs, its efficiency decreases drastically with respect to the GRT's one.

HSP-2 was not faster than the other planners in any domain, being always outperformed by FF. This was expected, since the two planners use the forward direction both for the construction of their heuristics and for traversing the state-space, however FF constructs its heuristic less times than HSP-2. Our impression is that the FF heuristic is also more informative and more accurate than the one of HSP-2. Concerning ALTALT, although it constructs its heuristic once, it did not manage to be faster than the others in any domain and this is (we believe) due to the problems that arise from the backward direction in which it traverses the state-space. So, this is an indication that in the case where opposite directions are used for the heuristic construction and the search phase, as GRT, ALTALT and HSPr do, it is preferable to use the backward direction for the heuristic construction and the forward direction for the search phase. This is why the problems that arise when constructing the heuristic backwards may be confronted more easily than the problems that arise when traversing the state-space backwards.

Domain analysis techniques, which occur in pre-processing phase, also play an important role. STAN, which is primarily based on these techniques, had many variations in its performance. In transportation domains, like the *logistics* and the *elevator* ones, where STAN exploits specialized heuristics, it was among the fastest planners. In the *gripper* domain, where STAN exploits its symmetry analysis, its performance was also excellent. In other domains, as for example the *freecell* or the *blocks*, it was not competitive due to its GRAPHPLAN basic architecture, which is not considered a fast technology any more.

FF also employed a domain analysis technique concerning goal ordering, which played an important role in the blocks problems. It would be very interesting to see the adaptation and the impact of this technique to other planners as well. As far as we know, HSP-2 and ALTALT are not using any domain analysis technique. GRT exploited only the domain enrichment technique in the *elevator* domain, however this technique is an integral part of its heuristic mechanism, in order to overcome some problems that arise from the backward heuristic construction.

An interesting observation concerns the performance of GRT in the bigger problems of the *logistics*, *freecell*, *gripper* and *puzzle* domains, where GRT exhibited better performance than in the smaller problems of the same domains, compared to the other planners. We believe that this is due to the fact that GRT constructs its heuristic once, while the repeated construction of the heuristics for the other planners is an inhibitory factor in the bigger problems.

The conclusions drawn above ignore a significant factor, which is the specific implementation, i.e. the approaches adopted by the various planners for "trivial" tasks, such as the computation of all the ground facts and actions of a problem or the computation of the applicable actions to a given state, the optimization of the code and of course potential "bugs". For example, in order to find the applicable actions to a state, GRT uses constraint satisfaction techniques to progressively instantiate the action schemas for each state, whereas most of the other planners exploit connectivity graphs between the facts of a problem and the pre-instantiated actions. Our experiments with GRT have shown that a significant portion of the processing time is spent in the determination of the applicable actions to a state. This is the reason why we have developed a parallel version of GRT, named PGRT (Vrakas et. al., 1999; 2000), which makes use of this observation and has been proved very efficient in all domains. However, it is in our future plans to develop a connectivity graph also in GRT and to compare it to the existent approach.

Differences that are due to the code optimization or potential "bugs" cannot be easily detected, but we believe that all the planners, both the oldest and the newest ones are well-optimized programs. In the future we would like to see theoretical comparisons between the computational complexities of the various techniques and algorithms, apart from their experimental evaluation that is usually adopted.

Concerning plan length, GRT produced better plans than the other planners in the *freecell* domain, in the *gripper* domain (along with other planners), in many *blocks* problems when a 3-action schemas representation was used and in some *logistics* problems. STAN exhibited the best behavior in most of the domains and we believe this is due to its GRAPHPLAN basic architecture, which always produces optimal parallel plans and, in many cases, sequential plans also. FF behaved well in the *logistics* and the *blocks* problems, with the 4-action schemas representation (in the latter case probably due to the goal ordering technique), however it produced lengthy plans in other domains, as the *freecell*, the *gripper* and the *puzzle* ones.

HSP-2 produced longer plans than GRT in many domains, as for example the *logistics*, the *freecell* and the *gripper* domains and the *blocks* one, when a 3-actions representation was used. This observation means that in these domains the related facts employed by the GRT heuristic proved more valuable than the forward and repeated reconstruction of the HSP-2 heuristic. Finally, ALTALT has not been distinguished for the quality of its plans in any domain.

Our general impression from the experiments is that there are specific domains that favor specific planners. So, what is important is to investigate the reasons for that. We are currently working in exploring the internal characteristics of each domain, classifying them into more general categories that share common features, and associate these features with specific heuristic search techniques. A first attempt for a domain classification can also be found in (Hoffmann, 2001).

An alternative view of the above problem concerns the way a domain is encoded. The same planner in the same domain may alter its performance when a different representation is adopted. We faced this problem with the *blocks-world*, with the 4- and 3-actions schemas domain representations, where the performance of GRT varies significantly, while the performance of other planners is also altered. We also faced this problem with the *elevator* and *movie* domains, which were the motivation for the development of the domain enrichment technique. Our conviction is that domain-independent planning is strongly domain-representation dependent.

Concerning GRT, we plan to extend it so as to handle more expressive domains, supporting most of the features of the PDDL language (types, quantifications, negations, disjunctions, etc). At this time we are working with an extension of GRT, which has the ability to take into account multiple criteria (i.e. solution time, resources, safety, profit etc.). We are also interested in incorporating domain analysis techniques, as they have been developed in STAN and DISCOPLAN, in order to take advantage of specialized methods for handling specific types of problems or sub-problems. Finally, we will investigate the possibility and the utility of combining domain independent planning techniques with domain dependent ones, without loosing the generality of the planning system.

## Acknowledgments

The authors would like to thank Thomas Eiter, the editor in charge for this paper, and the anonymous reviewers for their helpful comments. We would like also to thank Dimitris Vrakas for his careful reading and suggestions in the final version of the paper. Finally, we would like to thank the researchers of the planning community for making their planners available, and more specifically Blai Bonet, Hector Geffner, Joerg Hoffman, Bernhard Nebel, Derek Long, Maria Fox, Romeo Sanchez Nigenda, XuanLong Nguyen and Subbarao Kambhampati.